%% file: main.tex
\documentclass[letterpaper]{article} 
\usepackage{aaai2026}  
\usepackage{times}  
\usepackage{helvet}  
\usepackage{courier}  
\usepackage[hyphens]{url}  
\usepackage{graphicx} 
\usepackage{natbib}  
\usepackage{caption} 
\usepackage{algorithm}
\usepackage{algorithmic}

\usepackage{multirow}
\usepackage{color}
\usepackage{amsmath}
\usepackage{amsfonts}
\usepackage{bm}
\usepackage{booktabs}
\usepackage[table]{xcolor}

\usepackage{graphicx}
\usepackage{makecell}

\usepackage{placeins}
\definecolor{red}{rgb}{1,0.2,0.2}
\definecolor{or}{rgb}{1,0.5,0.25}
\definecolor{green}{rgb}{0, 1, 0}
\definecolor{bl}{rgb}{0, 0, 1}
\definecolor{brown}{rgb}{0.59, 0.3, 0}
\definecolor{cyan}{rgb}{0, 1, 1}
\definecolor{c_lowbest}{rgb}{1.0,1.0,0.9}
\definecolor{c_highbest}{rgb}{0.9,1.0,0.9}

\newcommand{\best}[1]  {{\textbf{#1}}}
\newcommand{\second}[1]  {{\underline{#1}}}

\usepackage{newfloat}
\usepackage{listings}
\DeclareCaptionStyle{ruled}{labelfont=normalfont,labelsep=colon,strut=off} 
\floatstyle{ruled}
\newfloat{listing}{tb}{lst}{}
\floatname{listing}{Listing}
\title{MoBGS: Motion Deblurring Dynamic 3D Gaussian Splatting \\ for Blurry Monocular Video}
\author{
    Minh-Quan Viet Bui\textsuperscript{\rm 1}\thanks{Co-first authors (equal contribution).},
    Jongmin Park \textsuperscript{\rm 1}\footnotemark[1], \\
    Juan Luis Gonzalez \textsuperscript{\rm 2},
    Jaeho Moon \textsuperscript{\rm 1},
    Jihyong Oh \textsuperscript{\rm 3}\equalcontrib,
    Munchurl Kim \textsuperscript{\rm 1}\equalcontrib
}
\affiliations{
    \textsuperscript{\rm 1}Korea Advanced Institute of Science and Technology (KAIST)\\
    \textsuperscript{\rm 2}Flawless AI\\
    \textsuperscript{\rm 3}Department of Imaging Science, GSAIM, Chung-Ang University\\

}

\usepackage{bibentry}

\begin{document}

\maketitle

\begin{abstract}
We present MoBGS, a novel motion deblurring 3D Gaussian Splatting (3DGS) framework capable of reconstructing sharp and high-quality novel spatio-temporal views from blurry monocular videos in an end-to-end manner. Existing dynamic novel view synthesis (NVS) methods are highly sensitive to motion blur in casually captured videos, resulting in significant degradation of rendering quality. While recent approaches address motion-blurred inputs for NVS, they primarily focus on static scene reconstruction and lack dedicated motion modeling for dynamic objects. To overcome these limitations, our MoBGS introduces a novel Blur-adaptive Latent Camera Estimation (BLCE) method using a proposed Blur-adaptive Neural Ordinary Differential Equation (ODE) solver for effective latent camera trajectory estimation, improving global camera motion deblurring.  In addition, we propose a Latent Camera-induced Exposure Estimation (LCEE) method to ensure consistent deblurring of both a global camera and local object motions. Extensive experiments on the Stereo Blur dataset and real-world blurry videos show that our MoBGS significantly outperforms the very recent methods, achieving state-of-the-art performance for dynamic NVS under motion blur.
\end{abstract}

\section{Introduction}
\label{sec:intro}
Novel View Synthesis (NVS) has shown significant advancements in recent years, with applications spanning Virtual Reality (VR), Augmented Reality (AR), and film production. While dynamic NVS methods \cite{pumarola2021d, li2021neural, li2022neural, wang2022fourier, weng2022humannerf, oswald2014generalized, collet2015high, broxton2020immersive, Gao-ICCV-DynNeRF, tretschk2021non, li2023dynibar, attal2023hyperreel, park2023temporal, shao2023tensor4d, fridovich2023k, yang2023deformable3dgs, Wu_2024_CVPR, bae2024per, wang2024shape, lei2024moscadynamicgaussianfusion, park2024splinegs, huang2023sc} have made substantial progress in reconstructing realistic scenes from monocular videos and multi-camera setups, their performance remains highly dependent on the quality of the given 2D observations. In particular, motion blur, which frequently occurs in casually captured videos due to fast-moving objects \cite{pan2016blind, zhang2020deblurring} or camera shake \cite{bahat2017non, zhang2018adversarial}, poses a major challenge. Since NVS techniques rely on the precise reconstruction of scene geometry and appearance, the loss of sharp details caused by blur can severely degrade the temporal consistency and rendering fidelity.

\begin{figure*}[t]
    \centering
    \includegraphics[width=\linewidth,keepaspectratio]{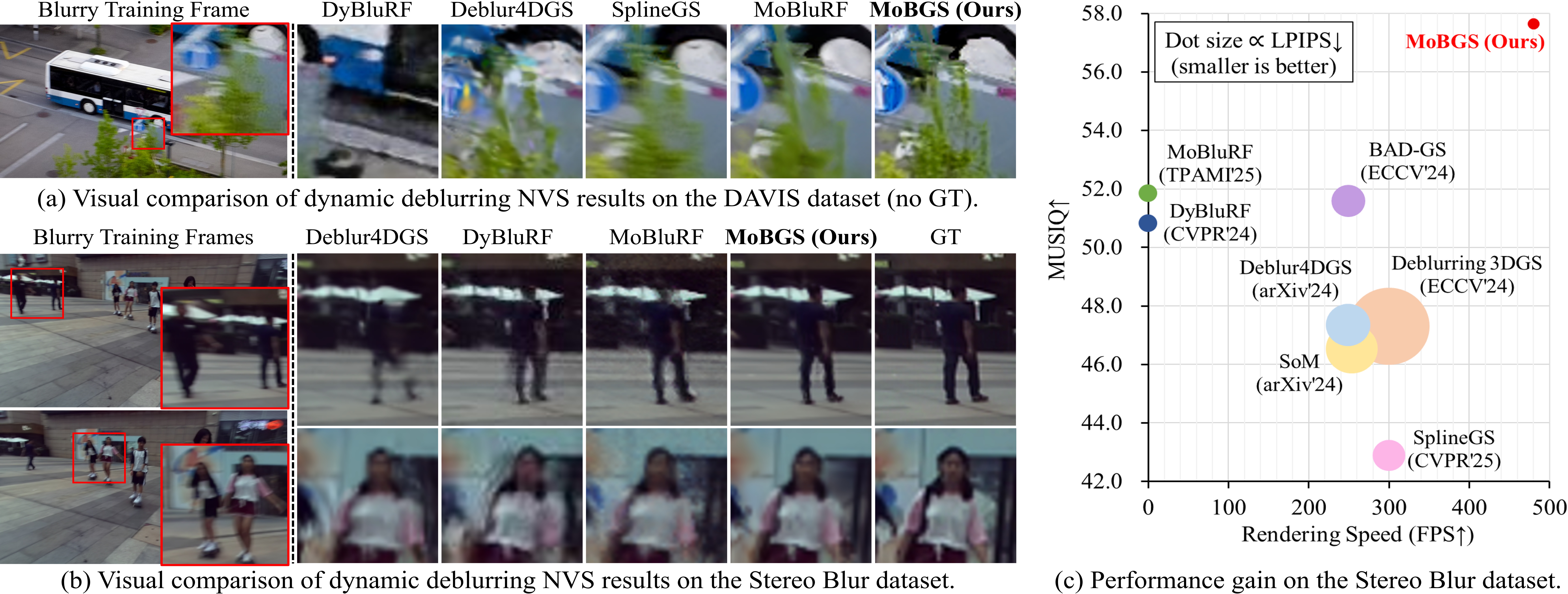}
   \captionof{figure}{Our MoBGS achieves state-of-the-art dynamic deblurring novel view synthesis on (a) real-world casually captured blurry monocular videos~\cite{ponttuset20182017davischallengevideo} and (b) synthesized blurry monocular videos \cite{sun2024dyblurf}, delivering (c) high perceptual quality (best LPIPS and MUSIQ scores) and fast rendering ($\sim$500 FPS). Each region is cropped and enlarged.}
   \label{fig:figure_page1}
\end{figure*}

To reconstruct sharp 3D scenes from blurry 2D observations, several deblurring NVS methods~\cite{ma2022deblur, wang2023bad, lee2023dp, lee2023exblurf, zhao2024bad, lee2024deblurring, lee2024crimgscontinuousrigidmotionaware, lee2024smurf, peng2025bags, sun2024dyblurf, bui2025moblurf, wu2024deblur4dgs} have been developed. These methods estimate camera trajectories during exposure, typically represented as latent camera poses, and generate a sequence of latent sharp images that are averaged to synthesize the blurred images, following the physical blur formation process \cite{zhao2024bad}. Most of these methods focus only on static scene deblurring, without considering that object motion also occurs during the same exposure time along with camera movement, making them less effective for dynamic scenes with complex object motions. 

A few recent methods~\cite{sun2024dyblurf, bui2025moblurf, wu2024deblur4dgs} have been explored for dynamic deblurring NVS from blurry monocular videos.
While these dynamic deblurring NVS methods represent significant progress, they still face critical limitations in modeling both global camera motion and local object motion blur. First, the prior methods~\cite{sun2024dyblurf, bui2025moblurf, wu2024deblur4dgs} for latent camera pose estimation lack guidance from the input blur degree. The blur degree of each input frame offers a valuable prior for accurately predicting the set of latent camera poses during exposure, as more severe blurriness typically corresponds to longer and more complex motion paths~\cite{chen2025image, fang2025parameterized}.
The second limitation is the inaccurate modeling of local motion blur, which results from motion being accumulated over a temporal interval.
Since object-induced blur accumulates over this interval, accurately estimating this interval is crucial for capturing its complex and spatially varying behavior~\cite{weng2023event, shang2023joint}.
However, recent methods do not explicitly incorporate this relationship when estimating the temporal span of motion integration or when modeling local motion blur.

To overcome these limitations and enable high-quality, sharp novel view synthesis from blurry monocular videos, we propose MoBGS, a novel \underline{Mo}tion de\underline{B}lurring method for NVS, based on dynamic 3D \underline{G}aussian \underline{S}platting~\cite{kerbl20233d}. Our MoBGS first accurately predicts latent camera poses for each blurry frame by leveraging the input frame's blur intensity as direct guidance using our Blur-adaptive Latent Camera Estimation (BLCE) method. Then, based on these estimated latent camera poses, our MoBGS estimates a latent exposure time for motion blur via our Latent Camera-induced Exposure Estimation (LCEE) method. The estimated latent exposure duration is then used to model motion blur from moving objects while ensuring the temporal consistency of global camera motion and local object motion blur.
Extensive quantitative and qualitative evaluations on the Stereo Blur dataset~\cite{sun2024dyblurf} and real-world videos~\cite{ponttuset20182017davischallengevideo} show that our MoBGS substantially outperforms recent state-of-the-art (SOTA) methods in dynamic deblurring NVS. Our main contributions are as follows:

\begin{itemize}
 \setlength\itemsep{0.05cm}
  \item We introduce a novel MoBGS framework for reconstructing high-quality spatio-temporal novel views from blurry monocular videos.
  \item A \textit{Blur-adaptive Latent Camera Estimation} (BLCE) method is proposed to accurately estimate latent camera trajectories, considering the blurriness of video frames via our novel Blur-adaptive Neural ODE.
  \item A \textit{Latent Camera-induced Exposure Estimation} (LCEE) method is proposed to estimate latent exposure time, ensuring consistent camera motion and object motion blur.
  \item Our MoBGS framework is extensively evaluated on the Stereo Blur dataset~\cite{sun2024dyblurf} and real-world blurry video sequences~\cite{ponttuset20182017davischallengevideo}, demonstrating \textit{significant} improvements over SOTA methods.
\end{itemize}

\section{Related Work}
\label{sec:related_work}

\noindent \textbf{Dynamic Novel View Synthesis}. Building on the success of Neural Radiance Field (NeRF)~\cite{mildenhall2020nerf} for static scenes, several methods have been developed for dynamic NVS. These include mapping observations to a shared canonical space~\cite{park2021nerfies, park2021hypernerf, song2023nerfplayer, liu2023robust}, estimating 3D scene flow~\cite{li2021neural, Gao-ICCV-DynNeRF, li2023dynibar, du2021neural}, and discretizing space-time volumes using grid-based architectures~\cite{cao2023hexplane, fridovich2023k, shao2023tensor4d, attal2023hyperreel}. More recently, 3DGS-based methods~\cite{kerbl20233d} have been proposed for efficient, high-fidelity dynamic NVS. These approaches model temporal deformation of Gaussians using per-frame attribute offsets~\cite{yang2023deformable3dgs}, grid-based encodings~\cite{wu20234d}, learned embeddings~\cite{bae2024per}, rigid transformations~\cite{wang2024shape}, motion decomposition \cite{kwak2025modecgs}, or spline-based trajectories~\cite{park2024splinegs}. While being effective under clean inputs, these methods fundamentally assume that each training frame is an instantaneous snapshot. As a result, they misinterpret motion blur as a feature of the scene. This leads to blur being baked into the 3D representation, resulting in blurry or ringing artifacts when trained on casually captured monocular videos.

\begin{figure*}
\centering
\includegraphics[width=\linewidth,keepaspectratio]{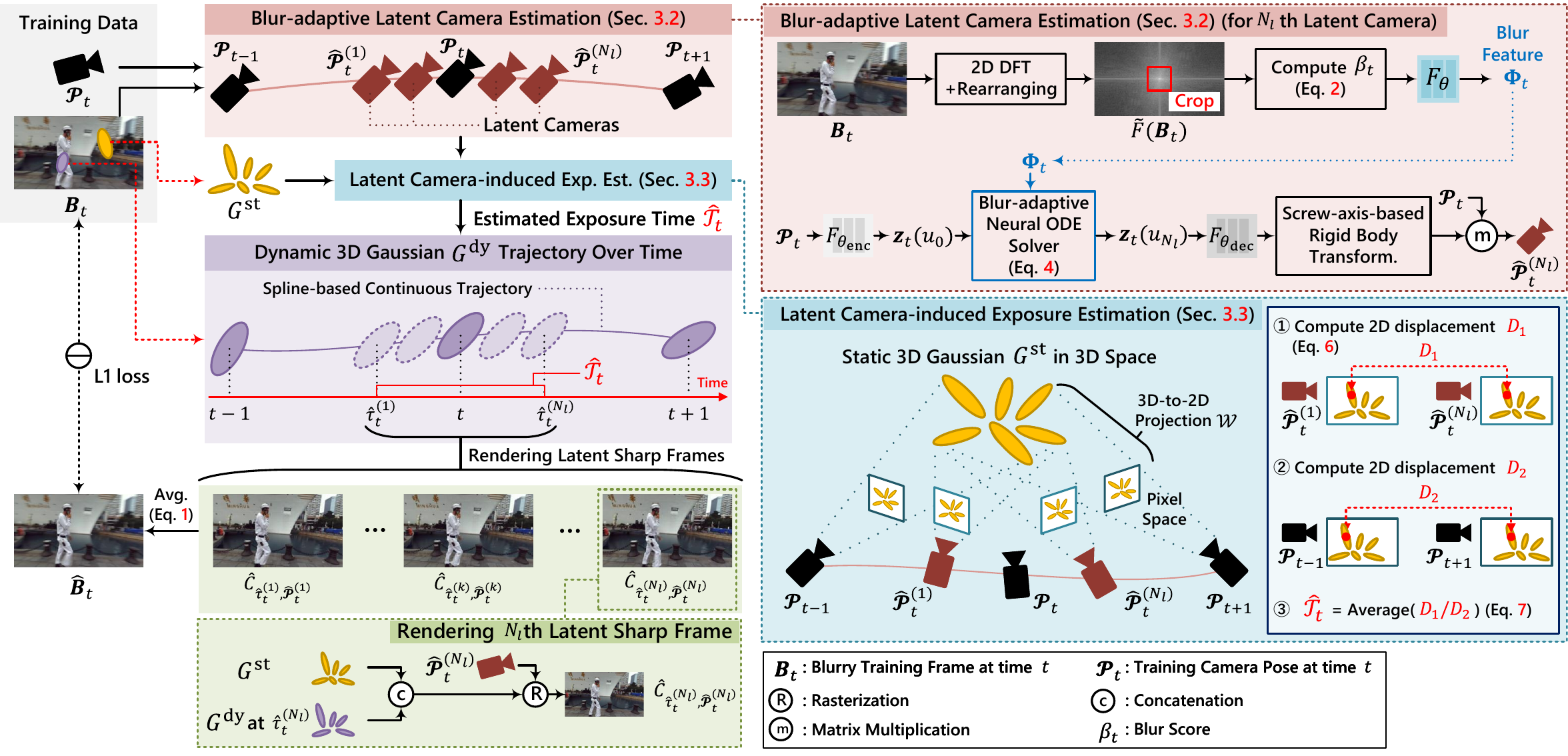}
\caption{\textbf{Overview of MoBGS.} MoBGS accurately models scene blurriness by jointly considering global camera and local object motion over the same exposure time. It first estimates latent camera poses for each blurry frame using the Blur-adaptive Latent Camera Estimation (BLCE) method. Then, leveraging these poses, it estimates the corresponding exposure time via the Latent Camera-induced Exposure Estimation (LCEE) method, ensuring a consistent blur modeling of local moving objects.}
\label{fig:overall_architecture}
\end{figure*}

\noindent \textbf{Deblurring Novel View Synthesis}. In deblurring NVS, early works like DeblurNeRF~\cite{ma2022deblur} and BAD-NeRF~\cite{wang2023bad} focused on modeling blur caused by camera motion, either through predicted blur kernels or estimated camera trajectories. These concepts were later adopted into more efficient 3DGS representation methods like Deblurring 3DGS~\cite{lee2024deblurring} and BAD-GS~\cite{zhao2024bad}. SMURF~\cite{lee2024smurf} was the first to use a Neural ODE~\cite{chen2018neural} to predict continuous latent camera trajectories. However, their per-pixel blur kernel estimation introduces inefficiency due to repeated computation. CRiM-GS~\cite{lee2024crimgscontinuousrigidmotionaware} improves efficiency by predicting global camera motion for all pixels but lacks the necessary prior information for accurate latent camera pose estimation. Importantly, none of these methods are designed to handle dynamic reconstruction.

The primary challenge in dynamic deblurring NVS has shifted to complex scenes where both camera and object motions contribute to the blur. Several methods have been proposed to tackle this problem~\cite{sun2024dyblurf, wu2024deblur4dgs, bui2025moblurf}, marking significant progress in the field. DyBluRF \cite{sun2024dyblurf} applies camera trajectory estimation from BAD-NeRF \cite{wang2023bad} on top of a dynamic NeRF, assuming fixed exposure time for object motion. MoBluRF \cite{bui2025moblurf} decomposes the 2D blur kernel into global camera and local object motion blur, enhancing sharp dynamic NeRF learning, although it omits exposure time estimation. Deblur4DGS~\cite{wu2024deblur4dgs} addresses dynamic deblurring NVS using 3DGS, incorporating regularization techniques and modeling object motion blur with a learnable exposure time. Despite these advancements, the prior methods still encounter difficulties in accurately modeling both global camera and local object blur components. First, their camera motion estimation lacks guidance from the blurred input frames, limiting adaptability. Second, their approaches to exposure time estimation overlook the simultaneous occurrence of global camera motion and local object motion blur. As a result, they struggle to robustly estimate the exposure time for modeling dynamic object motion blur.

\section{Proposed Method: MoBGS}
\label{sec:proposed_method}
\subsection{Overview of MoBGS}
Fig.~\ref{fig:overall_architecture} illustrates our MoBGS framework. Given a set of $N_f$ blurry monocular video frames $\{\bm{B}_t\}_{t=1}^{N_f}$ with corresponding camera poses $\{\bm{\mathcal{P}}_t\}_{t=1}^{N_f}$, where each $\bm{\mathcal{P}}_t \in \mathbb{R}^{4\times4}$, and a shared camera intrinsic matrix $\bm{K} \in \mathbb{R}^{3\times3}$, MoBGS aims to synthesize sharp novel views by accurately modeling scene blurriness caused by both global camera motion and local object motion. MoBGS is built on SplineGS~\cite{park2024splinegs}, which represents each scene with static $\{G^\text{st}_i\}_{i=1}^{n^\text{st}}$ and dynamic 3D Gaussians $\{G^\text{dy}_i\}_{i=1}^{n^\text{dy}}$, where the latter are deformed via splines to ensure smooth and continuous motion. More details are provided in \textit{Supplementary}.

Our blur modeling builds on two key insights: first, blur intensity serves as a valuable prior for predicting latent camera poses corresponding to global camera motion blur; second, both static backgrounds and moving objects undergo blurring during the same temporal interval in each frame.
Based on these insights, MoBGS systematically decouples global camera motion and local object motion blur components. To handle camera motion blur, the Blur-adaptive Latent Camera Estimation (BLCE) method (Sec.~\ref{sec:blce}) estimates a set of $N_l$ latent camera poses $\{\hat{\bm{\mathcal{P}}}^{(k)}_t\}_{k=1}^{N_l}$ for each blurry frame $\bm{B}_t$ by incorporating blur intensity as an additional input.
These poses are then used by the Latent Camera-induced Exposure Estimation (LCEE) method (Sec.~\ref{sec:lcee}) to compute the frame-wise \textit{latent} exposure time $\hat{\mathcal{T}}_t$, 
which represents a motion-integration interval that characterizes how much blur is accumulated within each frame.
To model spatially-varying blur from local object motion, we sample latent timestamps $\{\hat{\tau}^{(k)}_t\}_{k=1}^{N_l}$ within the estimated latent exposure interval $\hat{\mathcal{T}}_t$ and use them to predict corresponding object motion, enabling fine-grained motion blur modeling. 
Finally, we render each latent sharp frame at a sampled timestamp $\hat{\tau}^{(k)}_t$ and its corresponding latent camera pose $\hat{\bm{\mathcal{P}}}^{(k)}_t$, and denote it as $\hat{\bm{C}}_{\hat{\tau}^{(k)}_t, \hat{\bm{\mathcal{P}}}^{(k)}_t}$.
Following previous works~\cite{wang2023bad, lee2023dp, sun2024dyblurf, zhao2024bad, luthra2024deblur, wu2024deblur4dgs}, we formulate the final blurry frame by approximating continuous exposure with a discrete set of timestamps and averaging the corresponding latent sharp frames as:
\begin{equation}
    \hat{\bm{B}}_t = \frac{1}{N_{l}}\sum_{k=1}^{N_{l}} \hat{\bm{C}}_{\hat{\tau}^{(k)}_{t}, \hat{\bm{\mathcal{P}}}^{(k)}_{t}}, \quad \hat{\tau}^{(k)}_{t} = t + \hat{\mathcal{T}}_t \frac{k - \lceil \frac{N_l}{2} \rceil}{N_l},
    \label{eq:avg_latent}
\end{equation}
where $\lceil \cdot \rceil$ denotes the ceiling (round-up) operation. We refer readers to the \textit{Supplementary} for a detailed explanation of the blur modeling process underlying Eq.~\eqref{eq:avg_latent}.

\subsection{Blur-adaptive Latent Camera Estimation}
\label{sec:blce}
To address the blurriness caused by global camera motion, we propose the BLCE method which estimates latent camera poses guided by the blur intensity of each frame. To compute the blur intensity, we leverage the observation that blurry frames contain a higher proportion of low-frequency components due to the motion blur~\cite{liu2008image, shi2015just, shi2014discriminative, yan2016blind}. Based on this, we define the blur score as the ratio of the magnitudes within a low-frequency region to the total magnitudes in the frequency spectrum of the frame. Specifically, we apply a 2D Discrete Fourier Transform (DFT) to $\bm{B}_t$, and rearrange the resulting DFT coefficients to center the low-frequency components, producing the shifted DFT $\tilde{\mathcal{F}}(\bm{B}_t)$. The blur score $\beta_t$ is then given by:
\begin{equation}
    \beta_t = \frac{\sum_{\xi \in \Lambda} M_t(\xi)}{\sum_\xi M_t(\xi)}, \quad \text{where} \;\; M_t = |\tilde{\mathcal{F}}(\bm{B}_t)|,
    \label{eq:blur_score}
\end{equation}
where $\xi$ denotes a 2D frequency index and $\Lambda$ represents the center-cropped square region with a side length of $s$, which covers the low-frequency components in $\tilde{\mathcal{F}}(\bm{B}_t)$.  
 The blur feature $\bm{\Phi}_t$ is extracted from $\beta_t$ using a shallow MLP $F_\theta$ as:
\begin{equation}
    \bm{\Phi}_t = F_\theta(\phi(\beta_t)),
\end{equation}
where $\phi(\cdot)$ denotes the positional encoding~\cite{mildenhall2020nerf}.
Leveraging this blur feature $\bm{\Phi}_t$, we propose a novel Blur-adaptive Neural ODE solver, built upon Neural ODE~\cite{chen2018neural}, which predicts latent camera poses $\{\hat{\bm{\mathcal{P}}}_t^{(k)}\}_{k=1}^{N_l}$ along a smooth, temporally ordered trajectory.
We encode the initial latent feature $\textbf{z}_t(u_0)$ by feeding the camera pose $\bm{\mathcal{P}}_t$ into a shallow MLP $F_{\theta_\text{enc}}$, such that $\textbf{z}_t(u_0) = F_{\theta_\text{enc}}(\bm{\mathcal{P}}_t)$.
A sequence of $N_l$ latent vectors $\{\textbf{z}_t(u_k)\}_{k=1}^{N_l}$ is then computed from $\textbf{z}_t(u_0)$ as:
\begin{equation}
    \textbf{z}_t(u_k) = \textbf{z}_t(u_0) + \int_{u_0}^{u_k} f(\textbf{z}_t(u), u, \bm{\Phi}_t; \psi)\, du,
\end{equation} 
where $f$ is a shallow neural network parameterized by $\psi$, representing the derivative of the latent feature ${d\textbf{z}_t(u)}/{du}$ and $u$ denotes a temporal coordinate within the exposure duration. Unlike the existing methods~\cite{lee2024smurf, lee2024crimgscontinuousrigidmotionaware}, we inject the blur feature $\bm{\Phi}_t$ to $f$, enabling it to be guided by the blur intensity of each frame.
$\textbf{z}_t(u_k)$ is used to predict a screw axis $(\bm{\omega}^{(k)}_t; \bm{v}^{(k)}_t) \in \mathbb{R}^6$ via a decoder $F_{\theta_\text{dec}}$ as:
\begin{equation}
    (\bm{\omega}^{(k)}_t; \bm{v}^{(k)}_t) = F_{\theta_\text{dec}}(\textbf{z}_t(u_k)).
\end{equation}
Finally, we predict the latent camera pose $\hat{\bm{\mathcal{P}}}^{(k)}_t$ by applying a residual transformation $\Psi(\bm{\omega}^{(k)}_t, \bm{v}^{(k)}_t)$ to the initial camera pose \( \bm{\mathcal{P}}_t \), resulting in $\hat{\bm{\mathcal{P}}}_t^{(k)} = \bm{\mathcal{P}}_t \Psi(\bm{\omega}^{(k)}_t, \bm{v}^{(k)}_t)$, where $\Psi(\cdot)$ denotes the screw-axis-based rigid-body transformation introduced in Nerfies~\cite{park2021nerfies}.
By injecting blur intensity guidance, the BLCE method facilitates more stable learning of latent camera poses, leading to improved dynamic deblurring NVS performance across the entire frame, as demonstrated in Table~\ref{table:stereo_quantitative}, Table~\ref{table:stereo_quantitative_deblurring}, and Table~\ref{table:ablation_study_embed}.

\input{table/stereo_quantitative}

\begin{figure*}[t]
    \centering    \includegraphics[width=\linewidth,keepaspectratio]{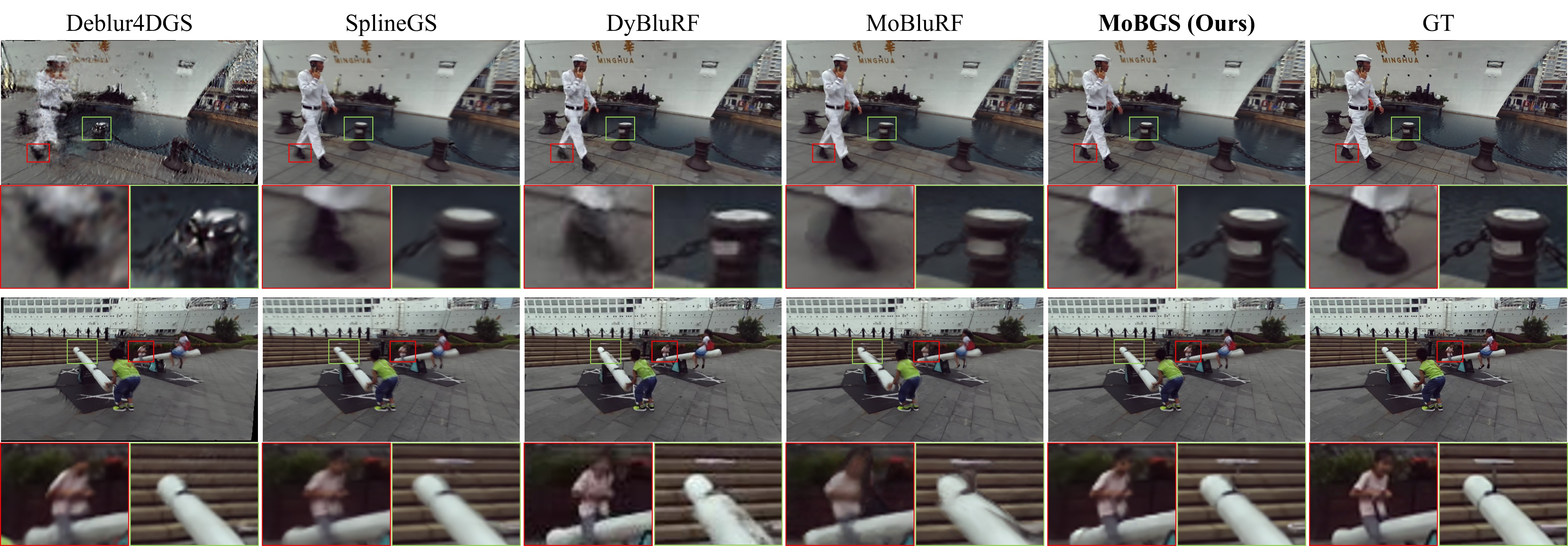}
    \caption{\textbf{Visual comparisons for dynamic deblurring novel view synthesis on the Stereo Blur dataset.}}
    \label{fig:qualitative_stereo}
\end{figure*}

\subsection{Latent Camera-induced Exposure Estimation}
\label{sec:lcee}
We propose the LCEE method to estimate the latent exposure time $\hat{\mathcal{T}}_t$ using prior knowledge from the latent camera trajectories predicted by BLCE (Sec.~\ref{sec:blce}). 
To avoid confusion, $\hat{\mathcal{T}}_t$ is not the physical exposure time; rather, it represents the learned temporal span over which motion blur accumulates, allowing the model to adapt to varying blur levels.
Given this interpretation of $\hat{\mathcal{T}}_t$, our key insight is that global camera motion and local object motion blur share the same exposure interval.
Motivated by this, we interpret $\hat{\mathcal{T}}_t$ as being proportional to the temporal interval between the first and last latent camera poses, $\hat{\bm{\mathcal{P}}}^{(1)}_{t}$ and $\hat{\bm{\mathcal{P}}}^{(N_l)}_{t}$.
As illustrated in the bottom-right blue box of Fig.~\ref{fig:overall_architecture}, we estimate the latent exposure time $\hat{\mathcal{T}}_t$ by comparing the magnitude of camera motion across two intervals. Based on the assumption that the consecutive training camera poses ${\bm{\mathcal{P}}}_{t-1}$ and ${\bm{\mathcal{P}}}_t$ are evenly separated by a single timestep, we compute $\hat{\mathcal{T}}_t$ as the ratio between the camera movement from $\hat{\bm{\mathcal{P}}}^{(1)}_{t}$ to $\hat{\bm{\mathcal{P}}}^{(N_l)}_{t}$ and that from ${\bm{\mathcal{P}}}_{t-1}$ to ${\bm{\mathcal{P}}}_{t+1}$. Since camera motion induces 2D displacements of static scene points on the image plane, we quantify the relative camera movement by measuring these projected displacements.
Let $\mathcal{W}(\bm{\mathcal{P}}, \bm{x})$ be a projection function that maps a 3D position $\bm{x}$ to a 2D pixel coordinate under a given camera pose $\bm{\mathcal{P}}$. Given the estimated latent camera poses $\{\hat{\bm{\mathcal{P}}}^{(k)}_{t}\}_{k=1}^{N_l}$ corresponding to each blurry frame $\bm{B}_t$, we select $\hat{\bm{\mathcal{P}}}^{(1)}_{t}$ and $\hat{\bm{\mathcal{P}}}^{(N_l)}_{t}$. As shown in Fig.~\ref{fig:overall_architecture}, we compute the 2D displacement of each static 3D Gaussian mean $\bm{\mu}^\text{st}_i$ between $\hat{\bm{\mathcal{P}}}^{(1)}_{t}$ and $\hat{\bm{\mathcal{P}}}^{(N_l)}_{t}$ as:
\begin{equation}
    \scalebox{0.9}{$
    D(\hat{\bm{\mathcal{P}}}^{(1)}_{t}, \hat{\bm{\mathcal{P}}}^{(N_l)}_{t}, \bm{\mu}^\text{st}_i) = \|\mathcal{W}(\hat{\bm{\mathcal{P}}}^{(1)}_{t}, \bm{\mu}^\text{st}_i) - \mathcal{W}(\hat{\bm{\mathcal{P}}}^{(N_l)}_{t}, \bm{\mu}^\text{st}_i)\|_2.
    \label{eq:2d_camera_latent}
    $}
\end{equation}
This projected displacement captures the shift of the static 3D Gaussian on the image plane caused by global camera motion between $\hat{\bm{\mathcal{P}}}^{(1)}_{t}$ and $\hat{\bm{\mathcal{P}}}^{(N_l)}_{t}$ during the exposure, representing the movement of the camera pair \( \hat{\bm{\mathcal{P}}}^{(1)}_{t} \) and \( \hat{\bm{\mathcal{P}}}^{(N_l)}_{t} \). Similarly, as shown in Fig. \ref{fig:overall_architecture}, we compute the 2D displacement between the two projections of the mean $\bm{\mu}^\text{st}_i$ for the \textit{training} (given) camera poses ${\bm{\mathcal{P}}}_{t-1}$ and ${\bm{\mathcal{P}}}_{t+1}$, representing the displacement over the interval from $t-1$ to $t+1$, denoted as $D({\bm{\mathcal{P}}}_{t-1}, {\bm{\mathcal{P}}}_{t+1}, \bm{\mu}^\text{st}_i)$.
To estimate the per-frame latent exposure time $\hat{\mathcal{T}}_t$, we compute the average ratio between the two 2D displacements as:
\begin{equation}
    \hat{\mathcal{T}}_t = \frac{2}{n^\text{st}} \sum_{i=1}^{n^\text{st}} \frac{D(\hat{\bm{\mathcal{P}}}^{(1)}_{t}, \hat{\bm{\mathcal{P}}}^{(N_l)}_{t}, \bm{\mu}^\text{st}_i) + \epsilon}{D({\bm{\mathcal{P}}}_{t-1}, {\bm{\mathcal{P}}}_{t+1}, \bm{\mu}^\text{st}_i) + \epsilon},
\label{eq:estimate_exposure}
\end{equation}
where $n^\text{st}$ is the number of static 3D Gaussians and $\epsilon$ is a smoothing term to avoid numerical instabilities. As shown in Eq.~\eqref{eq:avg_latent}, the estimated latent exposure time $\hat{\mathcal{T}}_t$ is used to compute the latent timestamps $\hat{\tau}^{(k)}_{t}$. Then, each latent sharp frame $\hat{\bm{C}}_{\hat{\tau}^{(k)}_{t}, \hat{\bm{\mathcal{P}}}^{(k)}_{t}}$ at each latent timestamp $\hat{\tau}^{(k)}_{t}$ is independently rendered using the static $\{G^\text{st}_i\}_{i=1}^{n^\text{st}}$ and dynamic 3D Gaussians $\{G^\text{dy}_i\}_{i=1}^{n^\text{dy}}$. Finally, the rendered blurry frame $\hat{\bm{B}}_t$ is obtained by averaging these latent sharp frames.
The LCEE effectively handles local object motion blur by leveraging the temporal alignment between global camera motion blur and local object motion blur.
We evaluate LCEE's effectiveness by comparing dynamic-region deblurring performance in Table~\ref{table:stereo_quantitative} and examining dynamic-region renderings under different exposure estimations in Table~\ref{table:ablation_study_lcee} and Fig.~\ref {fig:lcee_ablation}.

\subsection{Optimization}
We optimize our MoBGS using a photometric loss $\mathcal{L}_\text{rgb}$ and a depth loss $\mathcal{L}_\text{depth}$ defined as:
\begin{equation}
    \mathcal{L}_{\text{total}} = \lambda_{\text{rgb}} \mathcal{L}_{\text{rgb}} + \lambda_{\text{depth}} \mathcal{L}_{\text{depth}},
\label{eq:full_loss}
\end{equation}
where $\mathcal{L}_\text{rgb}$ is an L1 loss between the given blurry input frame $\bm{B}_t$ and our rendered blurry frame $\hat{\bm{B}}_t$. $\mathcal{L}_\text{depth}$ is an L1 loss between the rendered depth from rasterization and the ground-truth depth.

\section{Experiments}
\label{sec:experiments}

\noindent \textbf{Implementation Details.} We build our framework upon the widely adopted open-source 3D Gaussian Splatting (3DGS) codebase~\cite{kerbl20233d}.  For depth estimation and 2D tracking, we utilize the pre-trained models UniDepth \cite{piccinelli2024unidepth} and BootsTAP \cite{doersch2024bootstap}, respectively. Our model is trained over 10K iterations, with the LCEE applied after 2K iterations. For each blurry input frame, our BLCE module estimates $N_l = 9$ latent images. The blur score $\beta_t$ for the blur feature $\bm{\mathcal{B}}_t$ is computed using $s=20$. The loss coefficients are empirically found and set as follows: $\lambda_{\text{rgb}} = 1.0$, $\lambda_{\text{depth}} = 0.2$. We use $N_c=12$ control points for the spline-based motion modeling. Our MoBGS is trained and evaluated on a single NVIDIA RTX 3090Ti.

\noindent \textbf{Datasets.}
To assess the performance of our MoBGS and other SOTA methods, we utilize two datasets: the Stereo Blur dataset~\cite{sun2024dyblurf} and the DAVIS dataset~\cite{ponttuset20182017davischallengevideo}. The Stereo Blur dataset consists of stereo videos from six scenes with significant motion blur, each featuring a blurry left-view video and a corresponding sharp right-view video for evaluation. We follow the same training and testing splits as DyBluRF \cite{sun2024dyblurf}. The DAVIS dataset represents real-world scenarios with diverse scenes containing rapid object motions, resulting in natural motion blur. As ground-truth (GT) for NVS is not available, we adhere to the protocol used in \cite{park2024splinegs} by visualizing fixed-view, varying-time renderings to assess NVS quality.

\noindent \textbf{Metrics.} 
We evaluate the rendering quality of each method using LPIPS~\cite{zhang2018unreasonable}, MUSIQ~\cite{ke2021musiq}, and Peak Signal-to-Noise Ratio (PSNR). These metrics are computed across both the full image and dynamic regions. We further assess the temporal consistency across consecutive rendered frames using tOF~\cite{chu2020learning} score. Notably, increased blur can sometimes artificially elevate PSNR due to reduced sensitivity to small pixel misalignments, while perceptual metrics like LPIPS and MUSIQ better reflect human visual perception~\cite{zhang2018perceptual, barron2022mip, tucker2020fvs}. Therefore, we emphasize the importance of perceptual metrics, particularly LPIPS and MUSIQ, in evaluating deblurring performance.


\begin{figure}[!h]
    \centering
    \includegraphics[width=\linewidth,keepaspectratio]{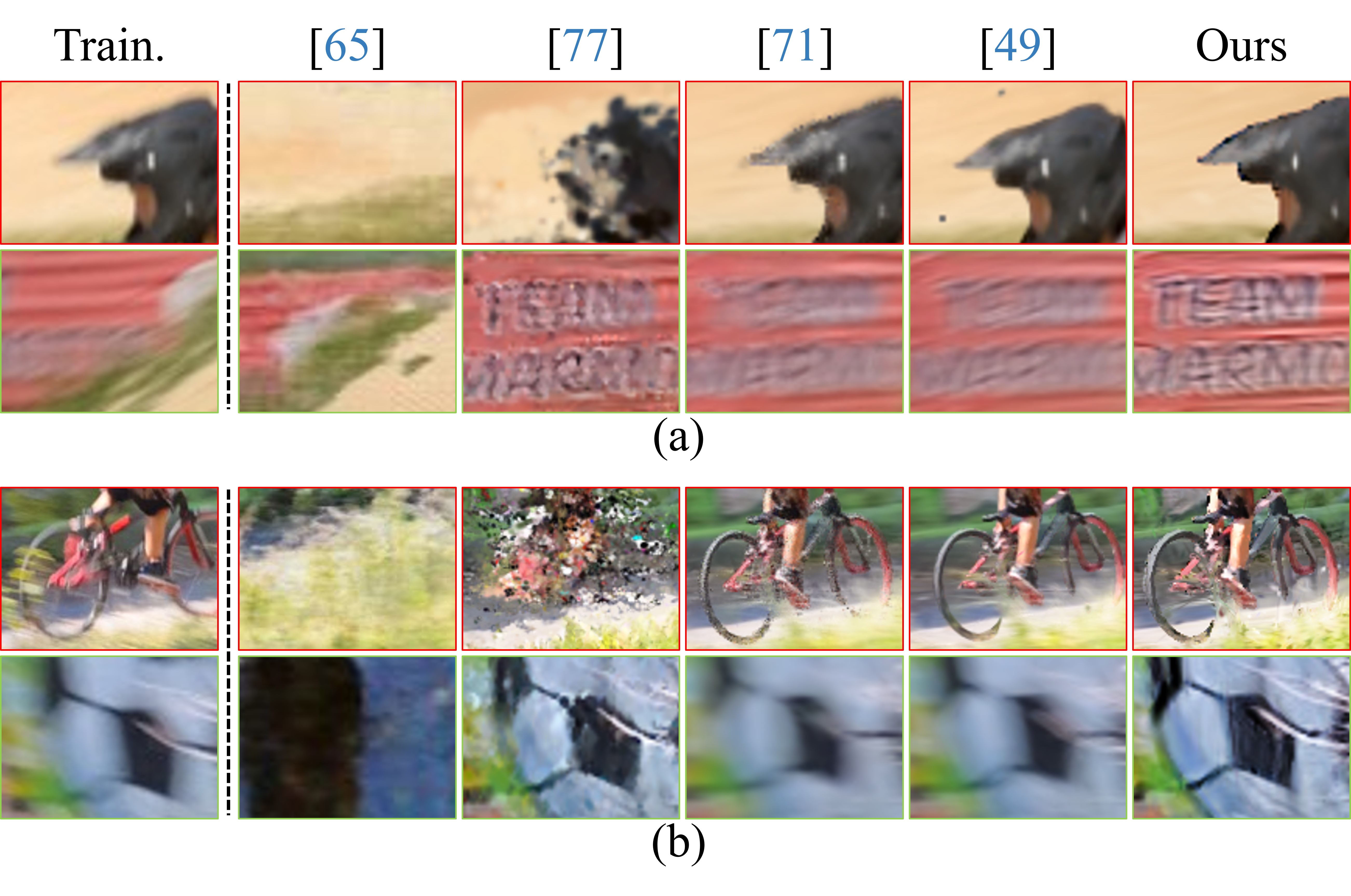}
    \caption{\textbf{Visual comparisons for dynamic deblurring novel view synthesis on the DAVIS dataset.} `Train.' refers to the training frame. The results of DyBluRF~\cite{sun2024dyblurf} exhibit significant misalignment, as its latent camera pose optimization is overfitted to the training poses. Full image results are provided in the \textit{Supplementary}.}
    \label{fig:qualitative_davis}
\end{figure}
\subsection{Comparison with State-of-the-Art Methods}
\label{sec:quanti}
\noindent \textbf{Dynamic Deblurring NVS.} The quantitative results for the joint evaluation of deblurring and dynamic NVS on the Stereo Blur dataset~\cite{sun2024dyblurf} are presented in Table \ref{table:stereo_quantitative}. As demonstrated, recent dynamic NVS methods \cite{yang2023deformable3dgs, wu2024deblur4dgs, bae2024per, wang2024shape, park2024splinegs} suffer from poor perceptual quality and low temporal consistency due to motion blur, as evidenced by higher (worse) LPIPS and tOF, alongside lower (worse) MUSIQ scores. This degradation stems from the lack of a dedicated deblurring module, which limits their abilities to effectively reconstruct sharp and temporally coherent novel views. Notably, our MoBGS achieves \textbf{2.7× higher perceptual scores} compared to the SOTA method, SplineGS. 

We then compare our results with cascade-based approaches which apply 2D deblurring networks as a preprocessing step. While the cascade approaches offer improvements over existing dynamic NVS methods, the combination of the two best methods (GShiftNet \cite{li2023simple} + SplineGS \cite{park2024splinegs}) still lags behind our MoBGS, particularly in structural accuracy, as reflected by the \textbf{1.3-2dB PSNR gaps} in both the full and dynamic regions. 

Table \ref{table:stereo_quantitative} also shows that the static deblurring NVS methods \cite{lee2024deblurring, zhao2024bad} fail to handle scene dynamics effectively. Notably, our MoBGS outperforms the latest state-of-the-art (SOTA) dynamic deblurring NVS method, MoBluRF~\cite{bui2025moblurf}, \textit{by very large margins in all metrics}, while achieving \textbf{4,800× faster} rendering speeds and \textbf{34× faster} training time. Figs. \ref{fig:figure_page1}, \ref{fig:qualitative_stereo} and \ref{fig:qualitative_davis} illustrate the qualitative results on both the Stereo Blur dataset~\cite{sun2024dyblurf} and the real-world videos from the DAVIS dataset~\cite{ponttuset20182017davischallengevideo}. Our visualizations show that our MoBGS yields \textit{significant} improvements in NVS quality, consistent with our quantitative results. More visualizations and demo videos are in the \textit{Supplementary}.

\noindent \textbf{Deblurring Effect.} Table~\ref{table:stereo_quantitative_deblurring} shows the deblurring effect on blurry training views. As expected, the dynamic NVS methods (SoM and SplineGS) tend to overfit to the blurriness of input frames, leading to low visual quality, as evidenced by poor LPIPS and MUSIQ scores. Consistent with our findings in Table \ref{table:stereo_quantitative} for the joint deblurring and dynamic NVS evaluation, our MoBGS outperforms cascade-based and dynamic deblurring NVS methods \textit{in all evaluation metrics}.
\input{table/stereo_deblurring_quantitative}

\input{table/ablation_embedding}
\vspace{-0.3cm}
\begin{figure}[h!]
    \centering
    \includegraphics[width=\linewidth,keepaspectratio]{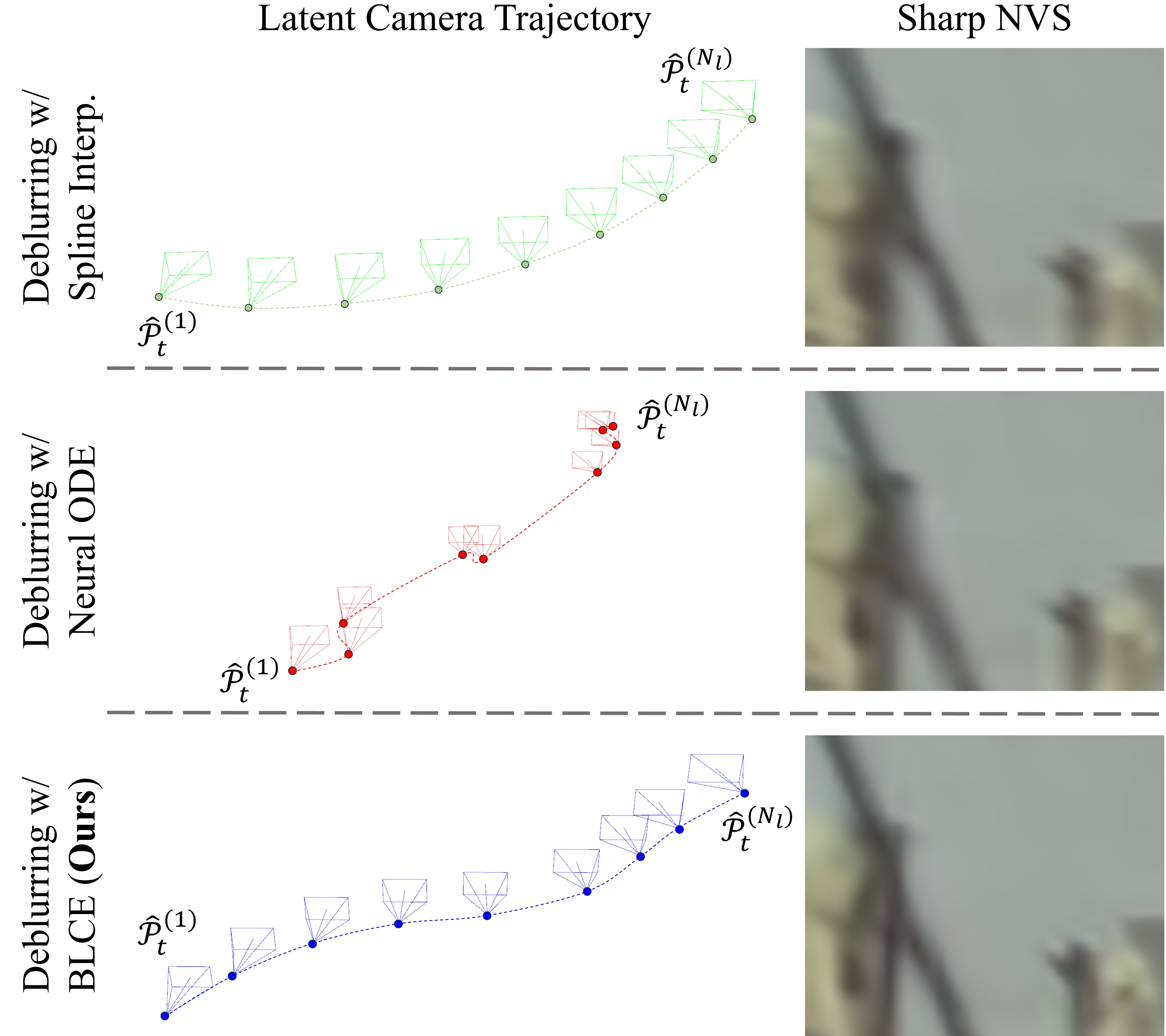}
    \caption{\textbf{Visual comparisons for BLCE ablation study.}}
    \label{fig:blce_ablation}
\end{figure}

\subsection{Ablation Study}
\label{sec:ablation}
\noindent \textbf{Blur-adaptive Latent Camera Estimation (BLCE).} As shown in Table~\ref{table:ablation_study_embed}, removing the deblurring mechanism (i.e., `No Deblurring module') results in a significant degradation of performance across all metrics, underscoring the negative impact of blurry training frames. These results highlight the effectiveness of our joint 3D reconstruction and deblurring approach for dynamic NVS. Additionally, we compare different latent camera estimation methods by replacing our BLCE with commonly used approaches from \cite{sun2024dyblurf, zhao2024bad}, which use spline interpolation (denoted as `Deblurring w/ Spline Interp.'), and from \cite{lee2024smurf, lee2024crimgscontinuousrigidmotionaware}, which use Neural ODE (denoted as `Deblurring w/ Neural ODE'). Table~\ref{table:ablation_study_embed} demonstrates that incorporating blur intensity guidance results in improvements across all perceptual metrics of the rendered images.

We present the latent camera trajectories and visual comparisons for the ablated models in Fig.~\ref{fig:blce_ablation}. As shown, `Deblurring w/ Spline Interp.' generates a smooth but oversimplified camera trajectory, which fails to capture fine details, leading to blurry reconstruction. On the other hand, `Deblurring w/ Neural ODE' produces jagged and inconsistent camera poses. In contrast, our BLCE with the novel Blur-adaptive Neural ODE solver produces a smooth camera trajectory while ensuring sharp reconstruction, which is consistent with our superior quantitative results in Table~\ref{table:ablation_study_embed}.

\input{table/ablation_lcee}

\begin{figure}[t]
    \centering
    \includegraphics[width=\linewidth,keepaspectratio]{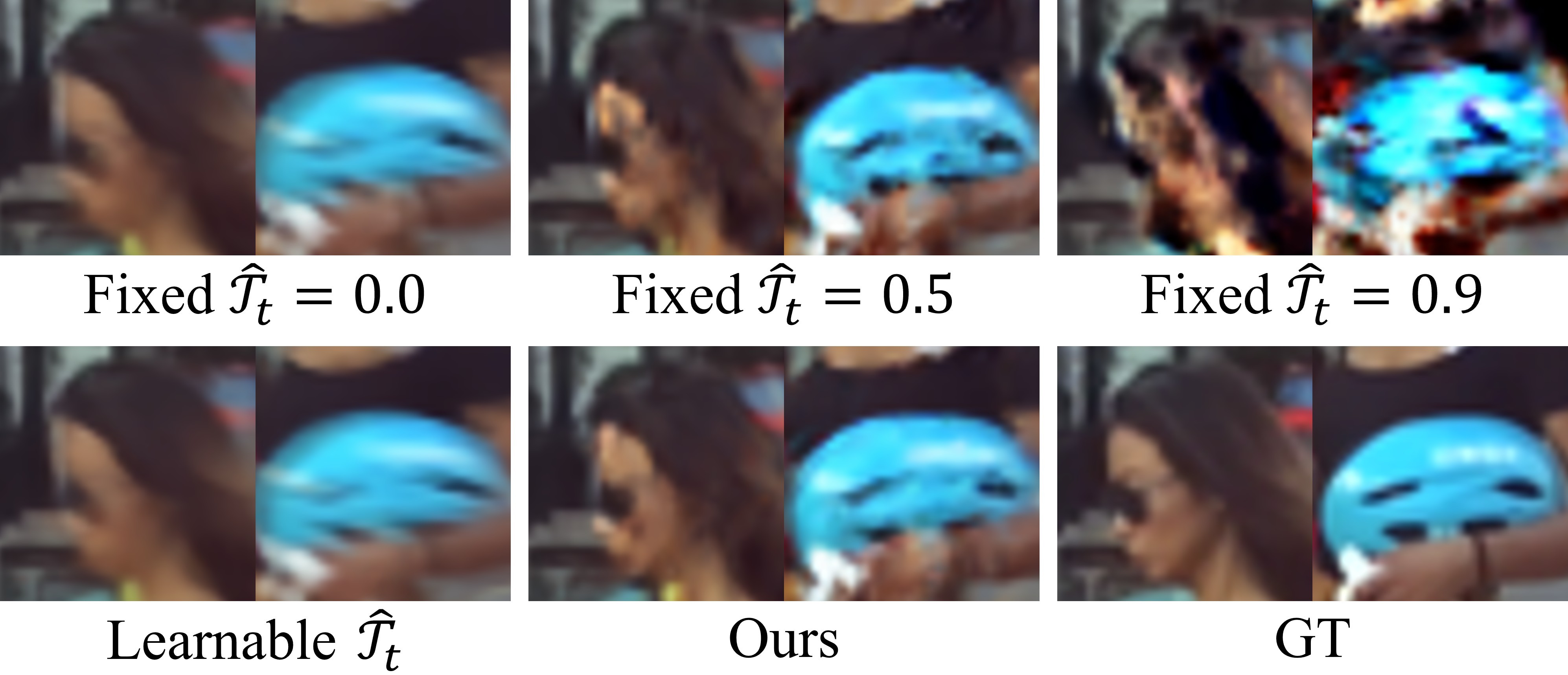}
    \caption{\textbf{Visual comparisons for LCEE ablation study.}}
    \label{fig:lcee_ablation}
\end{figure}

\begin{figure}[t]
    \centering
    \includegraphics[width=\linewidth,keepaspectratio]{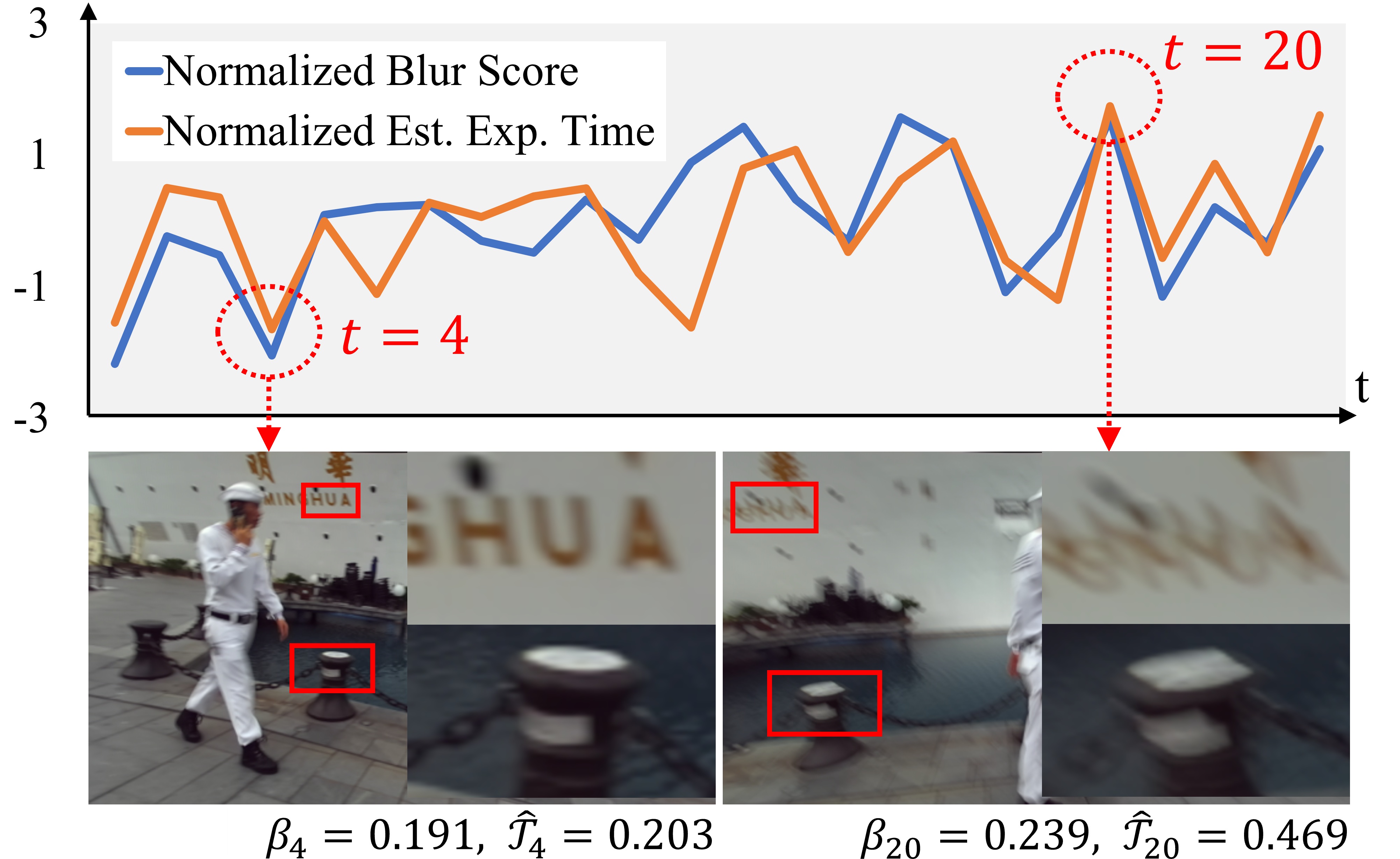}
    \caption{\textbf{Correlation between our estimated $\hat{\mathcal{T}}_t$ (Eq.~\eqref{eq:estimate_exposure}) and our blur score $\beta_t$ (Eq.~\eqref{eq:blur_score}).} For a clearer comparison of the trend, we normalize $\beta_t$ and $\hat{\mathcal{T}}_t$ separately using their respective means and standard deviations for the graph.}
    \label{fig:analysis_lcee}
\end{figure}

\noindent \textbf{Latent Camera-induced Exposure Estimation (LCEE).}
Unlike prior works \cite{bui2025moblurf, sun2024dyblurf, wu2024deblur4dgs}, our LCEE leverages the camera motion prior to estimate the latent exposure time, ensuring robust deblurring of dynamic regions. As shown in Fig.~\ref{fig:lcee_ablation}, the fixed latent exposure time setting (e.g., $\hat{\mathcal{T}}_t = 0.9$) introduces artifacts in dynamic regions, whereas setting $\hat{\mathcal{T}}_t = 0.0$ fails to account for any motion of dynamic objects, leading to blurry dynamic objects. The adverse effects of the above $\hat{\mathcal{T}}_t$ settings are also reflected in a worse perceptual score (LPIPS) and temporal inconsistencies (tOF) of \textit{dynamic regions} in Table \ref{table:ablation_study_lcee}. While the fixed latent exposure time $\hat{\mathcal{T}}_t = 0.5$ mitigates these artifacts, it remains suboptimal for handling varying levels of motion blur, requiring manual fine-tuning across different scenarios. Alternatively, treating $\hat{\mathcal{T}}_t$ as a learnable parameter without any constraints, as in \cite{wu2024deblur4dgs}, results in inconsistent deblurring between static and dynamic regions. This approach degrades the deblurring effect on moving objects and the temporal consistency, as reflected in the metrics of the `Learnable $\hat{\mathcal{T}}_t$' variant in Table \ref{table:ablation_study_lcee}, and is evidenced by the noisy artifacts in Fig.~\ref{fig:lcee_ablation}. 
 
We further evaluate the robustness of our estimated $\hat{\mathcal{T}}_t$ by plotting its normalized values alongside the normalization of blur score $\beta_t$ in Fig.~\ref{fig:analysis_lcee}. By analyzing the correlation between these values and the corresponding blurry inputs, we observe that $\hat{\mathcal{T}}_t$ effectively aligns with the degree of blurriness present in the input frames. This correlation demonstrates that our latent exposure time estimation is both reliable and adaptive to varying levels of motion blur, leading to robust deblurring performance across a wide range of scenarios.

\section{Conclusion}
We introduce MoBGS, a novel dynamic deblurring NVS method for blurry monocular videos. We propose the Blur-adaptive Latent Camera Estimation (BLCE) and Latent Camera-induced Exposure Estimation (LCEE) methods to accurately model scene blurriness. Extensive experiments on both real-world and synthetic blurry video datasets demonstrate that MoBGS significantly outperforms recent SOTA methods in dynamic deblurring NVS.

\section*{Acknowledgment}
This work was supported by Institute of Information and communications Technology Planning and Evaluation (IITP) grant funded by the Korean Government [Ministry of Science and ICT (Information and Communications Technology)] (Project Number: RS-2022-00144444, Project Title: Deep Learning Based Visual Representational Learning and Rendering of Static and Dynamic Scenes, 100\%).

\appendix
\maketitle
\section{Preliminaries}
\noindent \textbf{Motion Blur Formulation.} As in previous works \cite{wang2023bad, lee2023dp, sun2024dyblurf, zhao2024bad, luthra2024deblur, wu2024deblur4dgs}, a motion-blurred image $\bm{B} \in \mathbb{R}^{H\times W\times 3}$ can be expressed as the integration of latent sharp images $\bm{C}_{\varphi}\in \mathbb{R}^{H\times W\times 3}$ captured over an exposure time $\mathcal{T}$ as
\begin{equation}
    \bm{B} = g\int_{0}^{\mathcal{T}} \bm{C}_{\varphi}d\varphi,
\end{equation}
where $g$ is a normalization factor and $\varphi$ is an infinitesimal time interval. This process is commonly approximated by averaging $N_l$ latent sharp images, each denoted by $\bm{C}_{\tau^{(k)}}$, sampled at a discrete timestamp $\tau^{(k)}$ within  $\mathcal{T}$ as 
\begin{equation}
    \bm{B} \approx \frac{1}{N_{l}}\sum_{k=1}^{N_{l}} \bm{C}_{\tau^{(k)}}.
    \label{eq:avg_latent_preliminary}
\end{equation}

\noindent \textbf{3D Gaussian Splatting.} 3DGS~\cite{kerbl20233d} explicitly represents 3D scenes using anisotropic 3D Gaussians, each defined by a mean  $\bm{\mu} \in \mathbb{R}^3$ and a covariance matrix $\bm{\Sigma} \in \mathbb{R}^{3\times3}$ as
\begin{equation}
    \scalebox{0.9}{$
    G(\bm{x}) = e^{-\frac{1}{2}(\bm{x}-\bm{\mu})^\top \bm{\Sigma}^{-1} (\bm{x}-\bm{\mu})}, \;\; \text{where} \; \bm{\Sigma} = \bm{R}\bm{S}\bm{S}^{\top}\bm{R}^{\top},
    $}
\label{eq:Gaussians}
\end{equation}
where $\bm{S} \in \mathbb{R}^{3\times3}$ and $\bm{R} \in \mathbb{R}^{3\times3}$ denote the scaling and rotation matrices of the 3D Gaussian, respectively.
To project each 3D Gaussian onto 2D domain for rendering, a 2D covariance matrix $\bm{\Sigma}' \in \mathbb{R}^{2\times2}$ is computed as $\bm{\Sigma}' = \bm{JW\Sigma W^{\top}J^{\top}}$, where $\bm{J}$ is the Jacobian of the affine approximation of the projective transformation~\cite{zwicker2001surface} and $\bm{W}$ is a viewing transformation matrix.
The rendered color {$\bm{C}(\bm{u})$} at 2D pixel location {$\bm{u}$} is computed by blending $\mathcal{N}$ depth-ordered 3D Gaussians overlapped at $\bm{u}$ as
\begin{equation}
    \bm{C}(\bm{u}) = \textstyle \sum_{i\in\mathcal{N}}\bm{c}_i\alpha_i\prod^{i-1}_{j=1}(1-\alpha_j),
    \label{eq:alpha_blending}
\end{equation}
where $\bm{c}_i$ and $\alpha_i$ are the color and the density of the $i^{\text{th}}$ 3D Gaussian, respectively, computed using learnable spherical harmonic (SH) coefficients $\mathcal{C} \in \mathbb{R}^{(k)}$ and an opacity $\sigma \in \mathbb{R}$.

\noindent \textbf{Spline-based 3D Gaussian Deformation.} To model dynamic scenes, SplineGS~\cite{park2024splinegs} represents each scene with a combination of static 3D Gaussians $\{G^\text{st}_i\}_{i=1}^{n^\text{st}}$ for the stationary background and dynamic 3D Gaussians $\{G^\text{dy}_i\}_{i=1}^{n^\text{dy}}$ for moving objects. To represent the smooth and continuous deformation of each dynamic Gaussian over time, SplineGS~\cite{park2024splinegs} utilizes a set of $N_c$ learnable control points $\textbf{P} = \{\textbf{p}_j \in \mathbb{R}^3\}_{j=1}^{N_c}$ as an additional attribute, where the time-varying mean $\bm{\mu}(t)$ at time $t$ for each dynamic Gaussian is formulated as $\bm{\mu}(t) = S(t, \textbf{P})$ that is specifically expressed as 
\begin{equation}
\scalebox{0.9}{$
\begin{aligned}
    S(t, \textbf{P}) = (2t_r^3 - 3t_r^2 + 1)\textbf{p}_{\lfloor t_s \rfloor} + (t_r^3 - 2t_r^2 + t_r)\textbf{m}_{\lfloor t_s \rfloor} \\
    + (-2t_r^3 + 3t_r^2)\textbf{p}_{\lfloor t_s \rfloor +1} + (t_r^3 - t_r^2)\textbf{m}_{\lfloor t_s \rfloor +1},  \\
    t_r = t_s - \lfloor t_s \rfloor, \quad t_s = t_n(N_c-1), \quad t_n = t / (N_f-1),
\end{aligned}
$}
\label{eq:spline}
\end{equation}
where $\textbf{m}_j$ denotes the approximated tangent at the control point $\textbf{p}_j$, computed as $\textbf{m}_j = \frac{\textbf{p}_{j+1} - \textbf{p}_{j-1}}{2}$. $N_f$ is the total number of training video frames. We adopt this spline-based deformation for dynamic 3D Gaussians to accurately capture the motion trajectories of moving objects and to effectively model the blurriness across the exposure time.

\noindent \textbf{Neural ODE.} Neural ODE \cite{chen2018neural} is first proposed as an approach that interprets neural networks as the derivatives of an ODE system,
where ODE represents the dynamics inherent to the hidden
states. Specifically, Neural ODEs are utilized to represent parameterized, time-continuous dynamics in the latent space,
providing a unique solution given an initial value and numerical differential equation solvers. Given a time-dependent representation $\bm{h}_t$, Neural ODE approaches parameterize the continuous dynamics of $\bm{h}_t$ using a neural network $f$ as
\begin{equation}
    \frac{d\bm{h}_t(u)}{du} = f(\bm{h}_t,t;\psi),
\end{equation}
where $\psi$ is the network's parameters. From the initial representation at $t=0$, we can estimate a continuous trajectory of $\bm{h}$ by iteratively updating the representation.

\section{Datasets} 
The Stereo Blur dataset~\cite{sun2024dyblurf} comprises stereo videos of six scenes with significant motion blur, where each scene contains a blurry video from the left view and a corresponding sharp video from the right view. Each video is captured using a ZED stereo camera at 60 fps. To synthesize blurriness in the left camera, a video frame interpolation (VFI) method~\cite{niklaus2017video} is applied to increase the frame rate up to 480 fps, followed by frame averaging, which is used as a common protocol to obtain blurred frames \cite{su2017deep, nah2017deep}. For each scene, 24 frames are extracted from the original video, and camera parameters are estimated using COLMAP~\cite{schonberger2016structure} for fair comparison with other methods. The DAVIS dataset reflects real-world scenarios with diverse scenes featuring rapid object motions, resulting in significant natural real-world blurriness across frames. As reported in \cite{liu2023robust}, COLMAP~\cite{schonberger2016structure} fails to estimate reliable camera poses for this dataset. To address this, we adopt a more stable method proposed in SoM~\cite{wang2024shape} on camera parameter estimation for ours and all other SOTA methods to do fair performance comparisons.

\section{Demo Video}
We provide a demo video, \textit{MoBGS\_demo.mp4}, presenting a comparative analysis of MoBGS against SOTA methods on the Stereo Blur dataset~\cite{sun2024dyblurf}, as reported in Table \ref{table:stereo_quantitative}. This supplementary resource provides comprehensive visual evidence, further reinforcing the effectiveness of MoBGS in dynamic novel view synthesis under motion blur conditions.

\section{Additional Ablation Study}
\noindent \textbf{Ablation study on $\bm{N_l}$.} We conduct an ablation study on the number of latent sharp frames ($N_l$) in Table \ref{table:ablation_study_latent}. The evaluation metrics are obtained by testing on dynamic deblurring NVS using the Stereo Blur dataset. As observed in prior works \cite{wang2023bad, lee2023dp, zhao2024bad}, we observe that increasing $N_l$ enhances the deblurring effect, particularly improving perceptual metrics (LPIPS, MUSIQ) and temporal consistency (tOF). Our model stabilizes when $N_l \geq 9$, leading us to set $N_l=9$ as the final configuration for training efficiency.
\input{table/ablation_latent_rays}

\noindent \textbf{Ablation study on $\bm{s}$.} We conduct an ablation study on the value of blur feature side length ($s$) in Table \ref{table:ablation_study_s}. The metrics are the dynamic deblurring NVS results on the Stereo Blur dataset. When $s$ is too large (e.g., 100), $\beta_t$ captures less meaningful low-frequency content, degrading performance.
\input{table/ablation_s}

\section{Additional Results on DAVIS Dataset}
\begin{figure*}[t]
    \centering
    \includegraphics[width=1.0\linewidth,keepaspectratio]{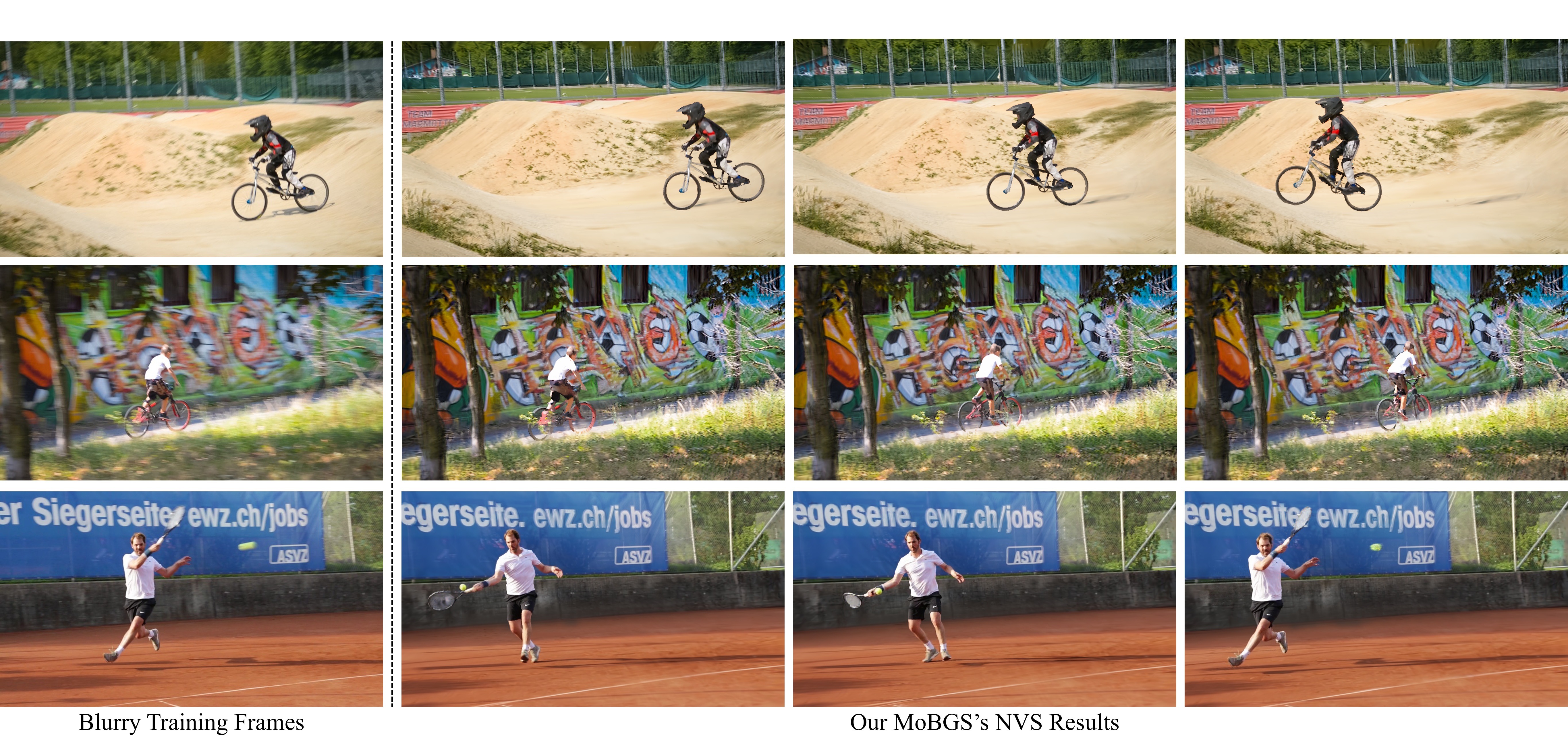}
    \caption{\textbf{Visual results of our MoBGS over time for dynamic deblurring novel view synthesis on the DAVIS dataset.}}
    \label{fig:qualitative_supple_davis}
\end{figure*}

\begin{figure*}[t]
    \centering
    \includegraphics[width=0.7\linewidth,keepaspectratio]{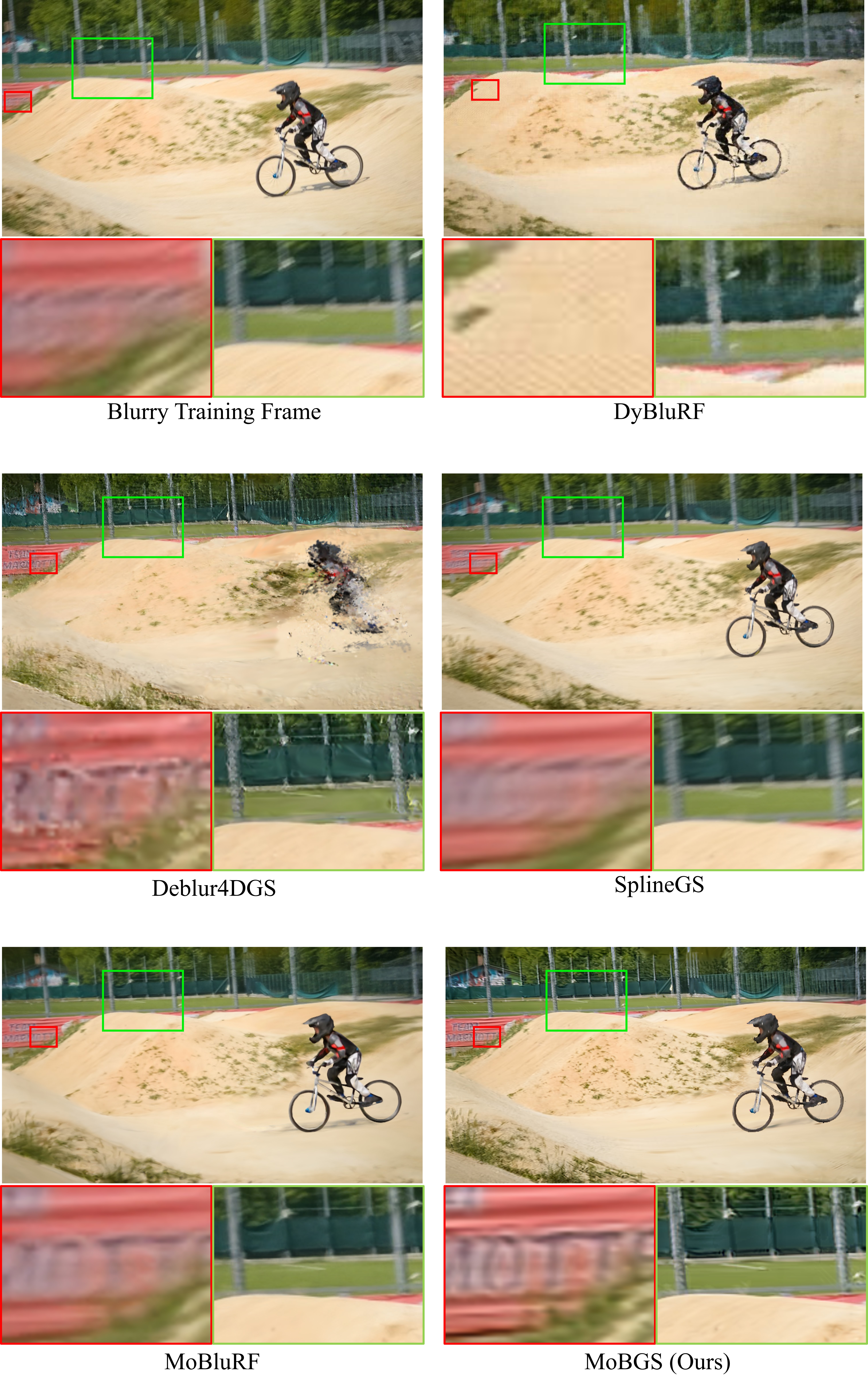}
    \caption{\textbf{Visual comparisons for dynamic deblurring novel view synthesis on the \textit{bmx-bumps} scene from the DAVIS dataset.}}
    \label{fig:qualitative_supple_bumps}
\end{figure*}

\begin{figure*}[t]
    \centering
    \includegraphics[width=0.7\linewidth,keepaspectratio]{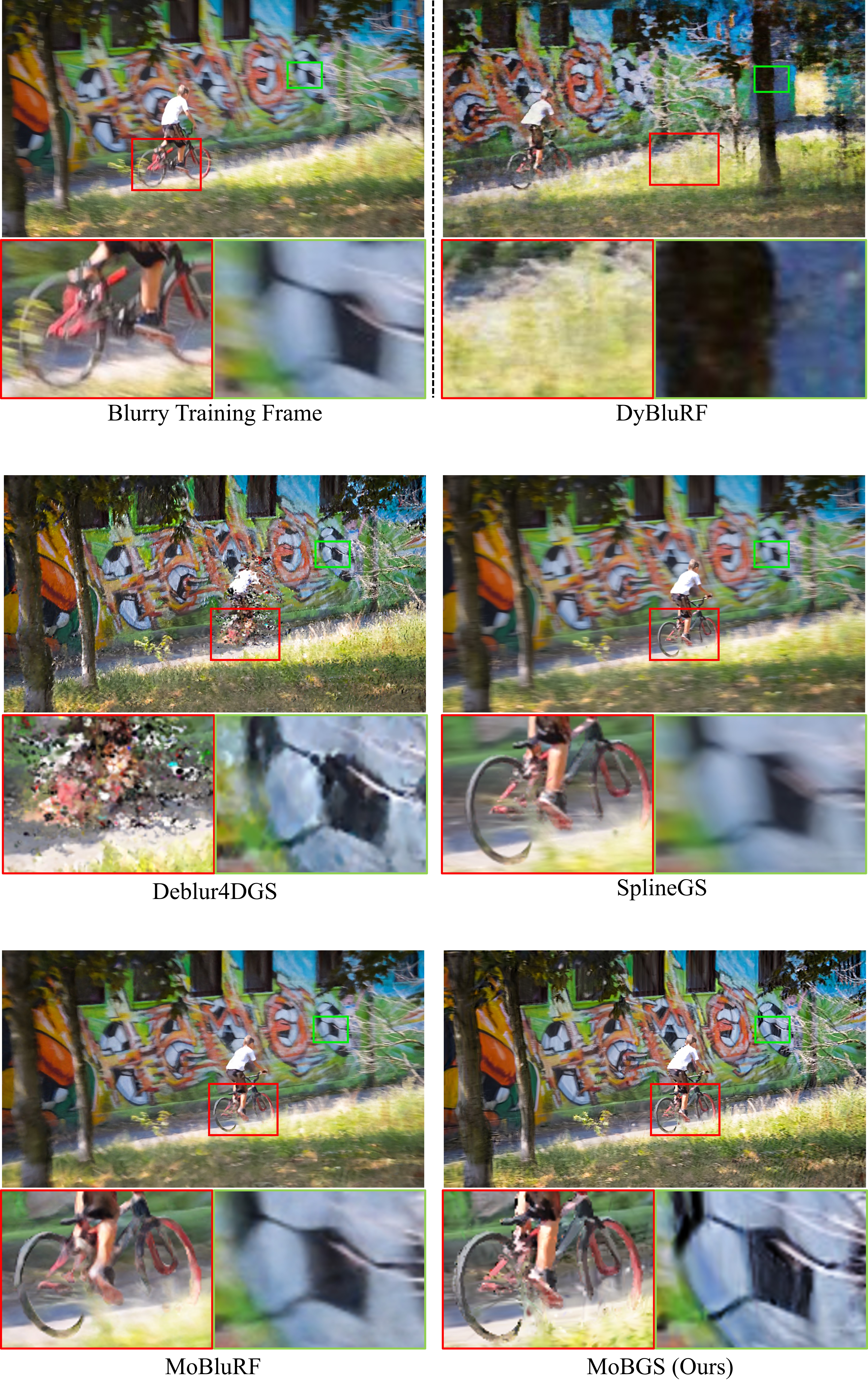}
    \caption{\textbf{Visual comparisons for dynamic deblurring novel view synthesis on the \textit{bmx-trees} scene from the DAVIS dataset.}}
    \label{fig:qualitative_supple_trees}
\end{figure*}

Fig. \ref{fig:qualitative_supple_davis} shows the spatio-temporal results of our MoBGS for dynamic deblurring NVS on the DAVIS dataset~\cite{ponttuset20182017davischallengevideo}.
Figs. \ref{fig:qualitative_supple_bumps} and \ref{fig:qualitative_supple_trees} present the full-region results of Fig.~\ref{fig:qualitative_davis} in the main paper.

\section{Per-scene results on Stereo Blur Dataset}
In Table \ref{table:perscene_lpips_nvs}, \ref{table:perscene_musiq_nvs}, \ref{table:perscene_tof_nvs}, \ref{table:perscene_psnr_nvs}, \ref{table:perscene_lpips_nvs_dynamic}, \ref{table:perscene_tof_nvs_dynamic}, \ref{table:perscene_psnr_nvs_dynamic}, we present the detailed quantitative results of all methods reported in Table \ref{table:stereo_quantitative}, broken down for each of the six scenes in the Stereo Blur dataset~\cite{sun2024dyblurf}. These tables provide per-scene performance metrics, including LPIPS, MUSIQ, tOF, and PSNR, enabling a comprehensive evaluation of the rendering quality, perceptual similarity, and temporal consistency across different methods. In addtion, the per-scene results of Table \ref{table:stereo_quantitative_deblurring} are detailed in Table \ref{table:perscene_lpips_deblurring}, \ref{table:perscene_musiq_deblurring}, \ref{table:perscene_tof_deblurring}, \ref{table:perscene_psnr_deblurring}.

\section{Limitation}
While MoBGS demonstrates strong performance on both synthetic and real-world blurry monocular videos, it currently operates within a per-scene optimization paradigm. Like other per-scene optimization approaches \cite{park2024splinegs, wang2024shape, yang2023deformable3dgs, wu20234d}, this design limits generalization to unseen scenes and necessitates re-optimization for new input sequences. Given the growing trend of feed-forward 3DGS models \cite{xu20254dgt, ren2024l4gm}, a promising future direction is to integrate the deblurring module into generalizable 3DGS methods to address the input degradation issue in generalizable Gaussian Splatting.

\input{table/perscene_lpips_nvs}
\input{table/perscene_musiq_nvs}
\input{table/perscene_tof_nvs}
\input{table/perscene_psnr_nvs}

\input{table/perscene_lpips_nvs_dynamic}
\input{table/perscene_tof_nvs_dynamic}
\input{table/perscene_psnr_nvs_dynamic}

\input{table/perscene_lpips_deblurring}
\input{table/perscene_musiq_deblurring}

\input{table/perscene_tof_deblurring}
\input{table/perscene_psnr_deblurring}

\clearpage
\clearpage

\bibliography{aaai2026}
\end{document}

%% file: table/stereo_quantitative.tex
\begin{table*}
\begin{center}
\caption{\textbf{Dynamic deblurring novel view synthesis evaluation on the Stereo Blur dataset.} {\textbf{Bold}} and {\underline{underline}} denote the best and second best performances, respectively. Per-scene results are provided in the \textit{Supplementary}.}
\setlength\tabcolsep{2pt} 
\renewcommand{\arraystretch}{1.1}
\scalebox{0.85}{
\begin{tabular}{ c | c | c | c | c | c}
\toprule
\multirow{2}{*}{} & \multirow{2}{*}{Methods} & Full image & Dynamic region & \multirow{2}{*}{FPS} & Train Time\\
& & \textbf{LPIPS$\downarrow$ / MUSIQ$\uparrow$ / tOF$\downarrow$ / PSNR$\uparrow$}  & \textbf{LPIPS$\downarrow$ / tOF$\downarrow$ / PSNR$\uparrow$} & & (hr)\\
\midrule
\multirow{4}{*}{\rotatebox[origin=c]{90}{{\parbox{1.5cm}{\centering Dynamic NVS}}}} 
& \multicolumn{1}{l|}{D3DGS (CVPR'24)~\cite{yang2023deformable3dgs}} & 0.446 / 34.33 / 6.972 / 15.91 & 0.542 / 8.215 / 13.81 & 57 & 0.30 \\
& \multicolumn{1}{l|}{4DGS (CVPR'24)~\cite{wu20234d}} & 0.270 / 39.74 / 5.400 / 18.53  & 0.367 / 7.574 / 14.95 & 193 & 0.25 \\
& \multicolumn{1}{l|}{SoM (arXiv'24)~\cite{wang2024shape}} & 0.219 / 46.52 / 2.395 / 22.92 & 0.272 / 2.876 / 17.18 & 254 & 1.2 \\
& \multicolumn{1}{l|}{SplineGS (CVPR'25)~\cite{park2024splinegs}} & 0.141 / 42.88 / 1.409 / 26.41 & 0.168 / 1.417 / \second{22.31} & 300 & 0.25 \\
\midrule
\multirow{4}{*}{\makecell{\rotatebox[origin=c]{90}{Cascade}}}
& \multicolumn{1}{l|}{Restormer (CVPR'22)~\cite{zamir2022restormer} + SoM} & 0.167 / 53.06 / 2.101 / 22.82 & 0.238 / 2.646 / 17.09 & 254 & 1.2 \\
& \multicolumn{1}{l|}{GShiftNet (CVPR'23)~\cite{li2023simple} + SoM} & 0.158 / 54.76 / 2.070 / 22.62 & 0.253 / 2.604 / 16.64 & 254 & 1.2 \\
& \multicolumn{1}{l|}{Restormer (CVPR'22)~\cite{zamir2022restormer} + SplineGS} & 0.081 / 54.01 / 0.911 / \second{26.83} & 0.118 / 1.249 / 22.26 & \second{336} & 0.25 \\
& \multicolumn{1}{l|}{GShiftNet (CVPR'23)~\cite{li2023simple} + SplineGS} & \second{0.074} / \second{55.29} / \second{0.748} / 26.54 & \second{0.108} / \second{1.174} / 22.06 & 329 & 0.25 \\
\midrule
\multirow{6}{*}{\makecell{\rotatebox[origin=c]{90}{\parbox{1.5cm}{\centering Deblurring NVS}}}}
& \multicolumn{1}{l|}{Deblurring 3DGS (ECCV'24)~\cite{lee2023exblurf}} & 0.347 / 47.30 / 3.342 / 15.76 & 0.466 / 6.559 / 12.63 & 300 & \best{0.18} \\
& \multicolumn{1}{l|}{BAD-GS (ECCV'24)~\cite{zhao2024bad}} & 0.146 / 51.58 / 1.551 / 21.43 & 0.330 / 4.207 / 15.18 & 250 & \second{0.2} \\
& \multicolumn{1}{l|}{DyBluRF (CVPR'24)~\cite{sun2024dyblurf}} & 0.079 / 50.82 / 0.889 / 25.62 & 0.158 / 1.367 / 19.41 & 0.2 & 51 \\
& \multicolumn{1}{l|}{Deblur4DGS (arXiv'24)~\cite{wu2024deblur4dgs}} & 0.191 / 47.33 / 1.880 / 23.37 & 0.299 / 3.248 / 17.53 & 250 & 2.4 \\
& \multicolumn{1}{l|}{MoBluRF (IEEE TPAMI)~\cite{bui2025moblurf}} & 0.078 / 51.84 / 0.816 / 25.69 & 0.155 / 1.456 / 20.63 & 0.1 & 50 \\
& \multicolumn{1}{l|}{\textbf{MoBGS (Ours)}} & \best{0.050} / \best{57.64} / \best{0.507} / \best{28.80} & \best{0.096} / \best{1.093} / \best{23.41}& \best{480} & 1.5 \\
\bottomrule
\end{tabular}
}
\label{table:stereo_quantitative}
\end{center}
\end{table*}

%% file: table/stereo_deblurring_quantitative.tex
\begin{table}
\begin{center}
\caption{\textbf{Deblurring effect on \textit{full images} on the Stereo Blur dataset.} Per-scene results are in the \textit{Supplementary}.}
\setlength\tabcolsep{6pt} 
\renewcommand{\arraystretch}{1.1}
\scalebox{0.9}{
\begin{tabular}{ c | c }
\toprule
Methods&\footnotesize \textbf{LPIPS$\downarrow$ / MUSIQ$\uparrow$ / tOF$\downarrow$ / PSNR$\uparrow$} \\
\midrule
\multicolumn{1}{l|}{SoM} & 0.207 / 46.55 / 2.268 / 24.53     \\ 
\multicolumn{1}{l|}{SplineGS} & 0.126 / 42.60 / 1.269 / 29.22   \\ 
\midrule
\multicolumn{1}{l|}{Restormer + SoM} & 0.151 / 52.60 / 1.905 / 24.70  \\  
\multicolumn{1}{l|}{GShiftNet + SoM} & 0.139 / 54.20 / 1.868 / 24.45   \\ 
\multicolumn{1}{l|}{Restormer + SplineGS} & 0.064 / 53.66 / 0.787 / \second{30.31}  \\ 
\multicolumn{1}{l|}{GShiftNet + SplineGS} & \second{0.058} / \second{55.24} / \second{0.641} / \second{30.31}  \\
\midrule
\multicolumn{1}{l|}{Deblurring 3DGS} & 0.220 / 50.88 / 2.262 / 20.47   \\
\multicolumn{1}{l|}{BAD-GS} & 0.119 / 53.39 / 1.527 / 25.27  \\
\multicolumn{1}{l|}{DyBluRF} & 0.064 / 52.16 / 0.740 / 29.15   \\
\multicolumn{1}{l|}{Deblur4DGS}  & 0.183 / 48.94 / 1.980 / 23.70  \\
\multicolumn{1}{l|}{MoBluRF} & 0.073 / 49.99 / 0.742 / 29.27  \\
\multicolumn{1}{l|}{\textbf{MoBGS (Ours)}} & \best{0.040 / \best{57.34} / \best{0.454}} / \best{30.96}  \\
\bottomrule
\end{tabular}
}
\label{table:stereo_quantitative_deblurring}
\end{center}
\end{table}

%% file: table/ablation_embedding.tex
\begin{table}
\centering
\centering
\caption{\textbf{Ablation on BLCE.} We evaluate dynamic deblurring NVS results of \textit{full image} on the Stereo Blur dataset.}
\setlength\tabcolsep{2.pt} 
\renewcommand{\arraystretch}{1.3}
\scalebox{0.8}{
\begin{tabular}{ l | c c c }
\bottomrule
\hline\noalign{\smallskip}
 \multicolumn{1}{c|}{Methods} & \textbf{LPIPS$\downarrow$}  & \textbf{MUSIQ$\uparrow$} & \textbf{tOF$\downarrow$} \\  
\bottomrule
  No Deblurring module (SplineGS) & {0.141} & {42.88} & 1.409 \\
  Deblurring w/ Spline Interp. (BAD-GS-like)  & {0.058} & {55.38} & 0.570 \\
  Deblurring w/ Neural ODE (CRiM-GS-like)& \second{0.057} & \second{55.49} & \second{0.536} \\
  \textbf{Deblurring w/ BLCE (Ours)} & \best{0.050} & \best{56.93} & \best{0.507} \\
\bottomrule
\hline\noalign{\smallskip}
\end{tabular}}
\label{table:ablation_study_embed}
\end{table}

%% file: table/ablation_lcee.tex
\begin{table}
\centering
\centering
\caption{\textbf{Ablation on LCEE.} We evaluate both dynamic deblurring NVS and deblurring effect results of \textit{dynamic region} on the Stereo Blur dataset.}
\setlength\tabcolsep{3.pt} 
\renewcommand{\arraystretch}{1.3}
\scalebox{0.8}{
\begin{tabular}{c|c | c}
\bottomrule
\hline\noalign{\smallskip}
\multirow{2}{*}{Methods} & Dynamic deblurring NVS & Deblurring effect \\
 & \textbf{LPIPS$\downarrow$ / tOF$\downarrow$ / PSNR$\uparrow$} & \textbf{LPIPS$\downarrow$ / tOF$\downarrow$ / PSNR$\uparrow$} \\ 
\bottomrule
\multicolumn{1}{l|}{Fixed $\hat{\mathcal{T}}_t=0.0$}
& 0.120 / \second{1.237} / 23.20 & 0.112 / \second{1.055} / 26.39 \\
\multicolumn{1}{l|}{Fixed $\hat{\mathcal{T}}_t=0.5$}
& \second{0.117} / 1.276 / 23.12 & \second{0.109} / 1.115 / 25.60 \\
\multicolumn{1}{l|}{Fixed $\hat{\mathcal{T}}_t=0.9$}
& 0.157 / 2.014 / 19.41 & 0.150 / 1.863 / 20.84 \\
\multicolumn{1}{l|}{Learnable $\hat{\mathcal{T}}_t$}  &
0.128 / 1.261 / \second{23.24} & 0.120 / 1.061 / \second{26.45} \\
\multicolumn{1}{l|}{\textbf{LCEE (Ours)}}                                                    
& \best{0.096} / \best{1.093} / \best{23.41} & \best{0.085} / \best{0.896} / \best{26.54} \\
\bottomrule
\hline\noalign{\smallskip}
\end{tabular}}
\label{table:ablation_study_lcee}
\end{table}

%% file: table/ablation_latent_rays.tex
\begin{table}
\centering
\centering
\caption{\textbf{Ablation on $\bm{N_l}$.} We evaluate dynamic deblurring NVS results of \textit{full image} on the Stereo Blur dataset.}
\setlength\tabcolsep{2.pt} 
\renewcommand{\arraystretch}{1.3}
\scalebox{0.8}{
\begin{tabular}{ l | c c c c }
\bottomrule
\hline\noalign{\smallskip}
 \multicolumn{1}{c|}{Methods} & \textbf{LPIPS$\downarrow$}   & \textbf{MUSIQ$\uparrow$} & \textbf{tOF$\downarrow$} & \textbf{PSNR$\uparrow$}\\  
\bottomrule
   $N_l=3$ (Train time: 0.8h)& {0.069} & 53.66 & 0.594 & \second{28.79} \\
    $N_l=5$ (Train time: 1.0h)& {0.055} & 56.48 & 0.526 & {28.78} \\
    $N_l=11$ (Train time: 2.0h)& \second{0.052} & \second{57.43} & \second{0.515} & {28.72} \\
  \textbf{$\bm{N_l=9}$ (Ours)}  (Train time: 1.5h)& \best{0.050} & \best{57.64} & \best{0.507} & \best{28.80} \\
\bottomrule
\hline\noalign{\smallskip}
\end{tabular}}
\label{table:ablation_study_latent}
\end{table}

%% file: table/ablation_s.tex
\begin{table}
\centering
\centering
\caption{\textbf{Ablation on $\bm{s}$.} We evaluate dynamic deblurring NVS results of \textit{full image} on the Stereo Blur dataset.}
\setlength\tabcolsep{2.pt} 
\renewcommand{\arraystretch}{1.3}
\scalebox{0.8}{
\begin{tabular}{ l | c c c c }
\bottomrule
\hline\noalign{\smallskip}
 \multicolumn{1}{c|}{Methods} & \textbf{LPIPS$\downarrow$}  & \textbf{MUSIQ$\uparrow$} & \textbf{tOF$\downarrow$} & \textbf{PSNR$\uparrow$}\\  
\bottomrule
   $s=10$ & 0.055 & \second{56.01} & \second{0.530} & \second{28.76} \\
    $s=50$ & \second{0.054} & {54.96} & 0.573 & 28.48 \\
    $s=100$ & {0.067} & 54.41 & 0.641 & {28.68} \\
  \textbf{$\bm{s=20}$ (Ours)} & \best{0.050} & \best{57.64} & \best{0.507} & \best{28.80} \\
\bottomrule
\hline\noalign{\smallskip}
\end{tabular}}
\label{table:ablation_study_s}
\end{table}

%% file: table/perscene_lpips_nvs.tex
\begin{table*}
\begin{center}
\caption{\textbf{Per-scene LPIPS of \textit{full region} for dynamic deblurring novel view synthesis evaluation on the Stereo Blur dataset.} }
\label{table:perscene_lpips_nvs}
\setlength\tabcolsep{6pt} 
\renewcommand{\arraystretch}{1.1}
\scalebox{0.85}{
\begin{tabular}{ c | c | c | c | c | c |c | c | c }
\toprule
 & Methods & Basketball & Children & Sailor & Seesaw & Skating & Street & \textbf{Average} \\
\midrule
\multirow{5}{*}{\rotatebox[origin=c]{90}{{\parbox{1.5cm}{\centering Dynamic NVS}}}}  
& \multicolumn{1}{l|}{D3DGS (CVPR'24)~\cite{yang2023deformable3dgs}} & 0.187	&0.490	&0.592	&0.228	&0.560	&0.617	&0.446 \\
& \multicolumn{1}{l|}{4DGS (CVPR'24)~\cite{wu20234d}} & 0.151&	0.315&	0.464	&0.131	&0.186&	0.375&	0.270 \\
& \multicolumn{1}{l|}{E-D3DGS (ECCV'24)~\cite{bae2024per}} & 0.179&	0.267&	0.359&	0.125	&0.166&	0.325&	0.237 \\
& \multicolumn{1}{l|}{SoM (arXiv'24)~\cite{wang2024shape}} & 0.158&	0.369	&0.306&	0.181	&0.163&	0.139	&0.219 \\
& \multicolumn{1}{l|}{SplineGS (CVPR'25)~\cite{park2024splinegs}}& 0.136	&0.243&	0.235	&0.079	&0.080	&0.072	&0.141 \\
\midrule
\multirow{6}{*}{\makecell{\rotatebox[origin=c]{90}{Cascade}}} 
& \multicolumn{1}{l|}{Restormer (CVPR'22)~\cite{zamir2022restormer} + E-D3DGS} & 0.144	&0.156&	0.299&	0.096&	0.144&	0.333	&0.195 \\
& \multicolumn{1}{l|}{GShiftNet (CVPR'23)~\cite{li2023simple} + E-D3DGS} & 0.135	&0.152	&0.305	&0.091&	0.125&	0.326	&0.189 \\
& \multicolumn{1}{l|}{Restormer (CVPR'22)~\cite{zamir2022restormer} + SoM} & 0.108	&0.248	&0.230&	0.145	&0.147&	0.127	&0.167 \\
& \multicolumn{1}{l|}{GShiftNet (CVPR'23)~\cite{li2023simple} + SoM} & 0.101	&0.230	&0.210&	0.128&	0.145&	0.133	&0.158\\
& \multicolumn{1}{l|}{Restormer (CVPR'22)~\cite{zamir2022restormer} + SplineGS} & 0.078&	0.133&	0.111&	0.046&	0.054	&0.064&	0.081\\
& \multicolumn{1}{l|}{GShiftNet (CVPR'23)~\cite{li2023simple} + SplineGS} & 0.067&	0.122	&0.107&	0.044	&\second{0.045	}&\second{0.056}&	\second{0.074} \\
\midrule
\multirow{6}{*}{\makecell{\rotatebox[origin=c]{90}{\parbox{1.5cm}{\centering Deblurring NVS}}}} 
& \multicolumn{1}{l|}{Deblurring 3DGS (ECCV'24)~\cite{lee2023exblurf}} & 0.207&	0.316	&0.496	&0.259&	0.430&	0.372&	0.347 \\
& \multicolumn{1}{l|}{BAD-GS (ECCV'24)~\cite{zhao2024bad}} & 0.085	&0.121	&0.190&	0.075	&0.130	&0.272&	0.146 \\
& \multicolumn{1}{l|}{DyBluRF (CVPR'24)~\cite{sun2024dyblurf}} & \second{0.050}&	\second{0.092}	&\second{0.115}	&0.075&	0.072	&0.068	&0.079\\
& \multicolumn{1}{l|}{Deblur4DGS (arXiv'24)~\cite{wu2024deblur4dgs}} & 0.134&	0.297	&0.352	&0.167&	0.118	&0.078	&0.191 \\
& \multicolumn{1}{l|}{MoBluRF (IEEE TPAMI)~\cite{bui2025moblurf}} & 0.066&	0.095	&\second{0.115}	&\second{0.042}&	0.080	&0.071	&0.078\\
& \multicolumn{1}{l|}{\textbf{MoBGS (Ours)}} & \best{0.042}&	\best{0.063}&	\best{0.090}	&\best{0.025}&	\best{0.033}&	\best{0.048}&	\best{0.050} \\
\bottomrule
\end{tabular}
}
\end{center}
\end{table*}

%% file: table/perscene_musiq_nvs.tex
\begin{table*}
\begin{center}
\caption{\textbf{Per-scene MUSIQ of \textit{full region} for dynamic deblurring novel view synthesis evaluation on the Stereo Blur dataset.} }
\setlength\tabcolsep{6pt} 
\renewcommand{\arraystretch}{1.1}
\scalebox{0.85}{
\begin{tabular}{ c | c | c | c | c | c |c | c | c }
\toprule
 & Methods & Basketball & Children & Sailor & Seesaw & Skating & Street & \textbf{Average} \\
\midrule
\multirow{5}{*}{\rotatebox[origin=c]{90}{{\parbox{1.5cm}{\centering Dynamic NVS}}}}  
& \multicolumn{1}{l|}{D3DGS (CVPR'24)~\cite{yang2023deformable3dgs}} & 35.60&29.92&34.26&41.31&30.56&34.35&34.33\\
& \multicolumn{1}{l|}{4DGS (CVPR'24)~\cite{wu20234d}} & 39.79&39.12&34.53&48.41&41.06&35.51&39.74 \\
& \multicolumn{1}{l|}{E-D3DGS (ECCV'24)~\cite{bae2024per}} & 38.59&35.94&33.08&44.92&42.41&43.43&39.73 \\
& \multicolumn{1}{l|}{SoM (arXiv'24)~\cite{wang2024shape}} & 32.69&48.11&44.97&44.47&56.04&52.84&46.52 \\
& \multicolumn{1}{l|}{SplineGS (CVPR'25)~\cite{park2024splinegs}}& 40.85&39.34&34.70&49.00&46.09&47.29&42.88 \\
\midrule
\multirow{6}{*}{\makecell{\rotatebox[origin=c]{90}{Cascade}}} 
& \multicolumn{1}{l|}{Restormer (CVPR'22)~\cite{zamir2022restormer} + E-D3DGS} & 52.43&52.12&46.16&50.69&41.17&43.38&47.66 \\
& \multicolumn{1}{l|}{GShiftNet (CVPR'23)~\cite{li2023simple} + E-D3DGS} & 55.86&50.81&50.20&53.50&47.35&42.03&49.96 \\
& \multicolumn{1}{l|}{Restormer (CVPR'22)~\cite{zamir2022restormer} + SoM} & 46.38&\best{61.33}&47.37&50.14&\second{57.57}&55.60&53.06 \\
& \multicolumn{1}{l|}{GShiftNet (CVPR'23)~\cite{li2023simple} + SoM} & 48.72&60.42&50.62&50.76&\best{59.14}&\best{58.92}&54.76\\
& \multicolumn{1}{l|}{Restormer (CVPR'22)~\cite{zamir2022restormer} + SplineGS} & 52.57&55.66&54.54&57.15&50.90&53.23&54.01\\
& \multicolumn{1}{l|}{GShiftNet (CVPR'23)~\cite{li2023simple} + SplineGS} & 54.11&54.96&52.29&\second{60.03}&54.33&56.01&\second{55.29}\\
\midrule
\multirow{6}{*}{\makecell{\rotatebox[origin=c]{90}{\parbox{1.5cm}{\centering Deblurring NVS}}}} 
& \multicolumn{1}{l|}{Deblurring 3DGS (ECCV'24)~\cite{lee2023exblurf}} & 47.31&49.03&40.77&58.04&44.01&44.67&47.30 \\
& \multicolumn{1}{l|}{BAD-GS (ECCV'24)~\cite{zhao2024bad}} & 49.95&57.93&\second{55.95}&54.64&43.69&47.34&51.58 \\
& \multicolumn{1}{l|}{DyBluRF (CVPR'24)~\cite{sun2024dyblurf}} & \second{54.34}&54.10&48.16&51.80&45.60&50.90&50.82\\
& \multicolumn{1}{l|}{Deblur4DGS (arXiv'24)~\cite{wu2024deblur4dgs}} & 43.19&41.64&48.80&47.68&48.61&54.09&47.33 \\
& \multicolumn{1}{l|}{MoBluRF (IEEE TPAMI)~\cite{bui2025moblurf}} & 51.03&55.37&51.81&57.48&45.87&49.463&51.84 \\
& \multicolumn{1}{l|}{\textbf{MoBGS (Ours)}} & \best{56.68}&\second{60.47}&\best{56.85}&\best{61.46}&53.36&\second{57.02}&\best{57.64} \\
\bottomrule
\end{tabular}
}
\label{table:perscene_musiq_nvs}
\end{center}
\end{table*}

%% file: table/perscene_tof_nvs.tex
\begin{table*}
\begin{center}
\caption{\textbf{Per-scene tOF of \textit{full region} for dynamic deblurring novel view synthesis evaluation on the Stereo Blur dataset.} }
\setlength\tabcolsep{6pt} 
\renewcommand{\arraystretch}{1.1}
\scalebox{0.85}{
\begin{tabular}{ c | c | c | c | c | c |c | c | c }
\toprule
 & Methods & Basketball & Children & Sailor & Seesaw & Skating & Street & \textbf{Average} \\
\midrule
\multirow{5}{*}{\rotatebox[origin=c]{90}{{\parbox{1.5cm}{\centering Dynamic NVS}}}}  
& \multicolumn{1}{l|}{D3DGS (CVPR'24)~\cite{yang2023deformable3dgs}} & 3.105&12.608&13.189&3.581&6.419&2.929&6.972\\
& \multicolumn{1}{l|}{4DGS (CVPR'24)~\cite{wu20234d}} & 2.912&12.137&10.357&2.634&2.030&2.329&5.400 \\
& \multicolumn{1}{l|}{E-D3DGS (ECCV'24)~\cite{bae2024per}} & 2.087&4.548&3.378&0.786&0.901&1.528&2.205 \\
& \multicolumn{1}{l|}{SoM (arXiv'24)~\cite{wang2024shape}} & 0.726&6.244&3.608&1.550&1.270&0.972&2.395 \\
& \multicolumn{1}{l|}{SplineGS (CVPR'25)~\cite{park2024splinegs}}& 0.654&4.606&1.697&0.491&0.521&0.482&1.409 \\
\midrule
\multirow{6}{*}{\makecell{\rotatebox[origin=c]{90}{Cascade}}} 
& \multicolumn{1}{l|}{Restormer (CVPR'22)~\cite{zamir2022restormer} + E-D3DGS} & 1.991&2.421&4.759&0.604&0.794&1.543&2.018 \\
& \multicolumn{1}{l|}{GShiftNet (CVPR'23)~\cite{li2023simple} + E-D3DGS} & 1.795&1.908&3.284&0.597&0.730&1.449&1.627 \\
& \multicolumn{1}{l|}{Restormer (CVPR'22)~\cite{zamir2022restormer} + SoM} & 0.734&5.052&3.367&1.298&1.190&0.965&2.101 \\
& \multicolumn{1}{l|}{GShiftNet (CVPR'23)~\cite{li2023simple} + SoM} & 0.708&5.132&2.855&1.471&1.225&1.031&2.070\\
& \multicolumn{1}{l|}{Restormer (CVPR'22)~\cite{zamir2022restormer} + SplineGS} & \second{0.556}&2.417&1.272&0.380&0.367&0.472&0.911\\
& \multicolumn{1}{l|}{GShiftNet (CVPR'23)~\cite{li2023simple} + SplineGS} & 0.638&1.531&\second{1.167}&\second{0.361}&\second{0.350}&\second{0.443}&\second{0.748}\\
\midrule
\multirow{6}{*}{\makecell{\rotatebox[origin=c]{90}{\parbox{1.5cm}{\centering Deblurring NVS}}}} 
& \multicolumn{1}{l|}{Deblurring 3DGS (ECCV'24)~\cite{lee2023exblurf}} & 1.103&2.907&6.828&1.703&5.957&1.552&3.342 \\
& \multicolumn{1}{l|}{BAD-GS (ECCV'24)~\cite{zhao2024bad}} & 0.645&1.486&2.943&0.509&1.620&2.103&1.551 \\
& \multicolumn{1}{l|}{DyBluRF (CVPR'24)~\cite{sun2024dyblurf}} & 0.676&1.460&1.423&0.662&0.592&0.518&0.889\\
& \multicolumn{1}{l|}{Deblur4DGS (arXiv'24)~\cite{wu2024deblur4dgs}} & 0.592&4.317&4.055&1.045&0.693&0.580&1.880 \\
& \multicolumn{1}{l|}{MoBluRF (IEEE TPAMI)~\cite{bui2025moblurf}} & 0.583&\second{1.312}&1.424&0.489&0.594&0.493&0.816 \\
& \multicolumn{1}{l|}{\textbf{MoBGS (Ours)}} &\best{0.335}&\best{0.818}&\best{0.927}&\best{0.231}&\best{0.322}&\best{0.410}&\best{0.507}\\
\bottomrule
\end{tabular}
}
\label{table:perscene_tof_nvs}
\end{center}
\end{table*}

%% file: table/perscene_psnr_nvs.tex
\begin{table*}
\begin{center}
\caption{\textbf{Per-scene PSNR of \textit{full region} for dynamic deblurring novel view synthesis evaluation on the Stereo Blur dataset.} }
\setlength\tabcolsep{6pt} 
\renewcommand{\arraystretch}{1.1}
\scalebox{0.85}{
\begin{tabular}{ c | c | c | c | c | c |c | c | c }
\toprule
 & Methods & Basketball & Children & Sailor & Seesaw & Skating & Street & \textbf{Average} \\
\midrule
\multirow{5}{*}{\rotatebox[origin=c]{90}{{\parbox{1.5cm}{\centering Dynamic NVS}}}}  
& \multicolumn{1}{l|}{D3DGS (CVPR'24)~\cite{yang2023deformable3dgs}} & 18.74&15.24&12.65&18.49&17.13&13.18&15.91\\  
& \multicolumn{1}{l|}{4DGS (CVPR'24)~\cite{wu20234d}} &20.75&18.31&14.77&21.65&21.44&14.27&18.53\\  
& \multicolumn{1}{l|}{E-D3DGS (ECCV'24)~\cite{bae2024per}} & 16.67&20.29&17.80&22.14&21.78&15.38&19.01\\  
& \multicolumn{1}{l|}{SoM (arXiv'24)~\cite{wang2024shape}} & 23.67&22.07&20.55&22.47&24.39&24.38&22.92\\  
& \multicolumn{1}{l|}{SplineGS (CVPR'25)~\cite{park2024splinegs}}& 24.98&25.02&24.37&28.09&29.09&\second{26.91}&26.41\\  
\midrule
\multirow{6}{*}{\makecell{\rotatebox[origin=c]{90}{Cascade}}} 
& \multicolumn{1}{l|}{Restormer (CVPR'22)~\cite{zamir2022restormer} + E-D3DGS} &16.41&19.67&17.57&22.12&21.55&14.97&18.72\\  
& \multicolumn{1}{l|}{GShiftNet (CVPR'23)~\cite{li2023simple} + E-D3DGS} & 16.33&19.76&17.27&22.16&21.82&15.79&18.85\\  
& \multicolumn{1}{l|}{Restormer (CVPR'22)~\cite{zamir2022restormer} + SoM} & 23.73&21.38&20.26&22.90&24.33&24.32&22.82\\  
& \multicolumn{1}{l|}{GShiftNet (CVPR'23)~\cite{li2023simple} + SoM} & 23.62&21.31&20.49&22.80&23.67&23.80&22.62\\  
& \multicolumn{1}{l|}{Restormer (CVPR'22)~\cite{zamir2022restormer} + SplineGS} & \second{25.55}&25.19&\second{24.84}&\second{29.51}&\second{29.53}&26.38&\second{26.83}\\  
& \multicolumn{1}{l|}{GShiftNet (CVPR'23)~\cite{li2023simple} + SplineGS} & 25.38&24.62&24.54&29.30&29.12&26.28&26.54\\  
\midrule
\multirow{6}{*}{\makecell{\rotatebox[origin=c]{90}{\parbox{1.5cm}{\centering Deblurring NVS}}}} 
& \multicolumn{1}{l|}{Deblurring 3DGS (ECCV'24)~\cite{lee2023exblurf}} & 17.63&16.75&12.63&18.76&15.31&13.49&15.76\\  
& \multicolumn{1}{l|}{BAD-GS (ECCV'24)~\cite{zhao2024bad}} & 24.21&21.71&19.19&23.13&22.41&17.95&21.43\\  
& \multicolumn{1}{l|}{DyBluRF (CVPR'24)~\cite{sun2024dyblurf}} & 25.28&\second{25.57}&23.50&24.56&27.94&26.88&25.62\\
& \multicolumn{1}{l|}{Deblur4DGS (arXiv'24)~\cite{wu2024deblur4dgs}} & 23.02&22.10&19.22&23.19&25.97&26.68&23.37\\ 
& \multicolumn{1}{l|}{MoBluRF (IEEE TPAMI)~\cite{bui2025moblurf}} & 24.28&23.78&24.08&28.14&27.64&26.21&25.69 \\
& \multicolumn{1}{l|}{\textbf{MoBGS (Ours)}} &\best{27.58}&\best{26.86}&\best{27.14}&\best{30.48}&\best{32.33}&\best{28.39}&\best{28.80} \\
\bottomrule
\end{tabular}
}
\label{table:perscene_psnr_nvs}
\end{center}
\end{table*}

%% file: table/perscene_lpips_nvs_dynamic.tex
\begin{table*}
\begin{center}
\caption{\textbf{Per-scene LPIPS of \textit{dynamic region} for dynamic deblurring novel view synthesis evaluation on the Stereo Blur dataset.} }
\setlength\tabcolsep{6pt} 
\renewcommand{\arraystretch}{1.1}
\scalebox{0.85}{
\begin{tabular}{ c | c | c | c | c | c |c | c | c }
\toprule
 & Methods & Basketball & Children & Sailor & Seesaw & Skating & Street & \textbf{Average} \\
\midrule
\multirow{5}{*}{\rotatebox[origin=c]{90}{{\parbox{1.5cm}{\centering Dynamic NVS}}}}  
& \multicolumn{1}{l|}{D3DGS (CVPR'24)~\cite{yang2023deformable3dgs}} & 0.306	&0.629	&0.691	&0.300	&0.637	&0.692	&0.542 \\
& \multicolumn{1}{l|}{4DGS (CVPR'24)~\cite{wu20234d}} & 0.266&	0.385&	0.561	&0.177	&0.300&	0.515&	0.367 \\
& \multicolumn{1}{l|}{E-D3DGS (ECCV'24)~\cite{bae2024per}} & 0.247&	0.296&	0.389&	0.151	&0.264&	0.393&	0.290 \\
& \multicolumn{1}{l|}{SoM (arXiv'24)~\cite{wang2024shape}} & 0.282&	0.338	&0.318&	0.228	&0.218&	0.246	&0.272 \\
& \multicolumn{1}{l|}{SplineGS (CVPR'25)~\cite{park2024splinegs}}& 0.135	&0.240&	0.271	&0.098	&0.089	&0.175	&0.168 \\
\midrule
\multirow{6}{*}{\makecell{\rotatebox[origin=c]{90}{Cascade}}} 
& \multicolumn{1}{l|}{Restormer (CVPR'22)~\cite{zamir2022restormer} + E-D3DGS} & 0.223	&0.186&	0.327&	0.128&	0.221&	0.396	&0.247 \\
& \multicolumn{1}{l|}{GShiftNet (CVPR'23)~\cite{li2023simple} + E-D3DGS} & 0.201	&0.193	&0.346	&0.114&	0.204&	0.393	&0.242 \\
& \multicolumn{1}{l|}{Restormer (CVPR'22)~\cite{zamir2022restormer} + SoM} & 0.177	&0.297	&0.295&	0.236	&0.201&	0.223	&0.238 \\
& \multicolumn{1}{l|}{GShiftNet (CVPR'23)~\cite{li2023simple} + SoM} & 0.217	&0.307	&0.264&	0.213&	0.236&	0.281	&0.253\\
& \multicolumn{1}{l|}{Restormer (CVPR'22)~\cite{zamir2022restormer} + SplineGS} & 0.101&	0.153&	0.192&	0.069&	0.061	&0.134&	0.118\\
& \multicolumn{1}{l|}{GShiftNet (CVPR'23)~\cite{li2023simple} + SplineGS} & \second{0.088}&	\second{0.149}	&\second{0.183}&	\second{0.059}	&\second{0.060}	&\second{0.110}&	\second{0.108} \\
\midrule
\multirow{6}{*}{\makecell{\rotatebox[origin=c]{90}{\parbox{1.5cm}{\centering Deblurring NVS}}}} 
& \multicolumn{1}{l|}{Deblurring 3DGS (ECCV'24)~\cite{lee2023exblurf}} & 0.334&	0.449	&0.584	&0.377&	0.547&	0.502&	0.466 \\
& \multicolumn{1}{l|}{BAD-GS (ECCV'24)~\cite{zhao2024bad}} & 0.260	&0.365	&0.425&	0.225	&0.274	&0.432&	0.330 \\
& \multicolumn{1}{l|}{DyBluRF (CVPR'24)~\cite{sun2024dyblurf}} & 0.156&	0.174	&0.224	&0.138&	0.113	&0.140	&0.158\\
& \multicolumn{1}{l|}{Deblur4DGS (arXiv'24)~\cite{wu2024deblur4dgs}} & 0.249&	0.509	&0.501	&0.175&	0.177	&0.181	&0.299 \\
& \multicolumn{1}{l|}{MoBluRF (IEEE TPAMI)~\cite{bui2025moblurf}} & 0.109&	0.203	&0.238	&0.106&	0.096	&0.180	&0.155 \\
& \multicolumn{1}{l|}{\textbf{MoBGS (Ours)}} & \best{0.071}&	\best{0.132}&	\best{0.168}	&\best{0.052}&	\best{0.043}&	\best{0.108}&	\best{0.096} \\
\bottomrule
\end{tabular}
}
\label{table:perscene_lpips_nvs_dynamic}
\end{center}
\end{table*}

%% file: table/perscene_tof_nvs_dynamic.tex
\begin{table*}
\begin{center}
\caption{\textbf{Per-scene tOF of \textit{dynamic region} for dynamic deblurring novel view synthesis evaluation on the Stereo Blur dataset.} }
\setlength\tabcolsep{6pt} 
\renewcommand{\arraystretch}{1.1}
\scalebox{0.85}{
\begin{tabular}{ c | c | c | c | c | c |c | c | c }
\toprule
 & Methods & Basketball & Children & Sailor & Seesaw & Skating & Street & \textbf{Average} \\
\midrule
\multirow{5}{*}{\rotatebox[origin=c]{90}{{\parbox{1.5cm}{\centering Dynamic NVS}}}}  
& \multicolumn{1}{l|}{D3DGS (CVPR'24)~\cite{yang2023deformable3dgs}} &4.998&12.149&14.954&3.643&6.821&6.723&8.215\\  
& \multicolumn{1}{l|}{4DGS (CVPR'24)~\cite{wu20234d}} & 5.358&14.090&13.900&3.104&2.289&6.705&7.574\\  
& \multicolumn{1}{l|}{E-D3DGS (ECCV'24)~\cite{bae2024per}} & 2.974&2.166&4.802&0.722&1.328&5.953&2.991\\  
& \multicolumn{1}{l|}{SoM (arXiv'24)~\cite{wang2024shape}} & 2.015&3.782&5.788&1.590&1.398&2.684&2.876\\  
& \multicolumn{1}{l|}{SplineGS (CVPR'25)~\cite{park2024splinegs}}& 1.219&2.156&2.456&0.512&0.507&1.651&1.417\\  
\midrule
\multirow{6}{*}{\makecell{\rotatebox[origin=c]{90}{Cascade}}} 
& \multicolumn{1}{l|}{Restormer (CVPR'22)~\cite{zamir2022restormer} + E-D3DGS} & 2.911&1.549&5.632&0.670&1.103&6.100&2.994\\  
& \multicolumn{1}{l|}{GShiftNet (CVPR'23)~\cite{li2023simple} + E-D3DGS} & 3.018&1.714&5.434&0.710&1.076&6.004&2.993\\  
& \multicolumn{1}{l|}{Restormer (CVPR'22)~\cite{zamir2022restormer} + SoM} & 1.739&3.330&5.610&1.419&1.151&2.629&2.646\\  
& \multicolumn{1}{l|}{GShiftNet (CVPR'23)~\cite{li2023simple} + SoM} & 2.015&3.505&4.319&1.593&1.340&2.850&2.604\\  
& \multicolumn{1}{l|}{Restormer (CVPR'22)~\cite{zamir2022restormer} + SplineGS} & 1.106&1.480&2.359&\second{0.471}&\second{0.458}&1.621&1.249\\  
& \multicolumn{1}{l|}{GShiftNet (CVPR'23)~\cite{li2023simple} + SplineGS} & \second{0.967}&\best{1.513}&\second{2.111}&0.479&0.500&\second{1.474}&\second{1.174}\\ 
\midrule
\multirow{6}{*}{\makecell{\rotatebox[origin=c]{90}{\parbox{1.5cm}{\centering Deblurring NVS}}}} 
& \multicolumn{1}{l|}{Deblurring 3DGS (ECCV'24)~\cite{lee2023exblurf}} &3.248&3.652&15.906&1.750&7.840&6.955&6.559\\  
& \multicolumn{1}{l|}{BAD-GS (ECCV'24)~\cite{zhao2024bad}} & 2.951&3.351&7.872&1.509&2.807&6.751&4.207\\  
& \multicolumn{1}{l|}{DyBluRF (CVPR'24)~\cite{sun2024dyblurf}} & 1.271&1.588&2.398&0.721&0.620&1.604&1.367\\  
& \multicolumn{1}{l|}{Deblur4DGS (arXiv'24)~\cite{wu2024deblur4dgs}} &2.119&6.517&6.903&0.884&1.130&1.935&3.248\\
& \multicolumn{1}{l|}{MoBluRF (IEEE TPAMI)~\cite{bui2025moblurf}} & 1.150&1.764&2.714&0.827&0.525&1.757&1.456 \\
& \multicolumn{1}{l|}{\textbf{MoBGS (Ours)}} & \best{0.881}&\second{1.521}&\best{1.980}&\best{0.414}&\best{0.367}&\best{1.395}&\best{1.093} \\
\bottomrule
\end{tabular}
}
\label{table:perscene_tof_nvs_dynamic}
\end{center}
\end{table*}

%% file: table/perscene_psnr_nvs_dynamic.tex
\begin{table*}
\begin{center}
\caption{\textbf{Per-scene PSNR of \textit{dynamic region} for dynamic deblurring novel view synthesis evaluation on the Stereo Blur dataset.} }
\setlength\tabcolsep{6pt} 
\renewcommand{\arraystretch}{1.1}
\scalebox{0.85}{
\begin{tabular}{ c | c | c | c | c | c |c | c | c }
\toprule
 & Methods & Basketball & Children & Sailor & Seesaw & Skating & Street & \textbf{Average} \\
\midrule
\multirow{5}{*}{\rotatebox[origin=c]{90}{{\parbox{1.5cm}{\centering Dynamic NVS}}}}  
& \multicolumn{1}{l|}{D3DGS (CVPR'24)~\cite{yang2023deformable3dgs}} & 17.47&12.04&10.16&15.82&13.69&13.67&13.81\\  
& \multicolumn{1}{l|}{4DGS (CVPR'24)~\cite{wu20234d}} & 17.47&14.22&11.07&18.03&15.35&13.54&14.95\\  
& \multicolumn{1}{l|}{E-D3DGS (ECCV'24)~\cite{bae2024per}} & 17.10&17.93&15.67&20.78&15.62&14.37&16.91\\  
& \multicolumn{1}{l|}{SoM (arXiv'24)~\cite{wang2024shape}} & 19.62&16.42&14.34&16.57&17.67&18.45&17.18\\  
& \multicolumn{1}{l|}{SplineGS (CVPR'25)~\cite{park2024splinegs}}& 22.82&20.81&\second{19.44}&24.60&\second{23.79}&\second{22.42}&\second{22.31}\\  
\midrule
\multirow{6}{*}{\makecell{\rotatebox[origin=c]{90}{Cascade}}} 
& \multicolumn{1}{l|}{Restormer (CVPR'22)~\cite{zamir2022restormer} + E-D3DGS} & 16.94&17.74&15.27&20.62&15.39&13.52&16.58\\  
& \multicolumn{1}{l|}{GShiftNet (CVPR'23)~\cite{li2023simple} + E-D3DGS} &16.54&17.66&14.44&20.36&15.73&14.00&16.46\\  
& \multicolumn{1}{l|}{Restormer (CVPR'22)~\cite{zamir2022restormer} + SoM} & 20.35&15.81&13.82&16.65&17.62&18.29&17.09\\  
& \multicolumn{1}{l|}{GShiftNet (CVPR'23)~\cite{li2023simple} + SoM} & 19.60&14.93&14.08&16.96&16.95&17.29&16.64\\  
& \multicolumn{1}{l|}{Restormer (CVPR'22)~\cite{zamir2022restormer} + SplineGS} &22.72&\best{21.52}&19.34&\second{24.67}&23.46&21.87&22.26\\  
& \multicolumn{1}{l|}{GShiftNet (CVPR'23)~\cite{li2023simple} + SplineGS} & 22.26&\second{21.35}&19.23&24.66&22.98&21.90&22.06\\  
\midrule
\multirow{6}{*}{\makecell{\rotatebox[origin=c]{90}{\parbox{1.5cm}{\centering Deblurring NVS}}}} 
& \multicolumn{1}{l|}{Deblurring 3DGS (ECCV'24)~\cite{lee2023exblurf}} & 16.09&12.47&10.16&14.74&11.70&10.65&12.63\\  
& \multicolumn{1}{l|}{BAD-GS (ECCV'24)~\cite{zhao2024bad}} & 19.63&12.89&12.59&15.69&16.35&13.92&15.18\\  
& \multicolumn{1}{l|}{DyBluRF (CVPR'24)~\cite{sun2024dyblurf}} &21.44&19.66&17.18&17.39&20.70&20.06&19.41\\  
& \multicolumn{1}{l|}{Deblur4DGS (arXiv'24)~\cite{wu2024deblur4dgs}} & 19.29&15.74&14.48&18.14&18.80&18.76&17.53 \\
& \multicolumn{1}{l|}{MoBluRF (IEEE TPAMI)~\cite{bui2025moblurf}} & \second{23.38}&15.52&18.88&21.58&22.51&21.89&20.63 \\
& \multicolumn{1}{l|}{\textbf{MoBGS (Ours)}} &\best{24.61}&20.92&\best{20.93}&\best{24.89}&\best{26.53}&\best{22.57}&\best{23.41} \\
\bottomrule
\end{tabular}
}
\label{table:perscene_psnr_nvs_dynamic}
\end{center}
\end{table*}

%% file: table/perscene_lpips_deblurring.tex
\begin{table*}
\begin{center}
\caption{\textbf{Per-scene LPIPS of \textit{full region} for deblurring effect evaluation on the Stereo Blur dataset.} }
\setlength\tabcolsep{6pt} 
\renewcommand{\arraystretch}{1.1}
\scalebox{0.85}{
\begin{tabular}{ c | c | c | c | c |c | c | c }
\toprule
  Methods & Basketball & Children & Sailor & Seesaw & Skating & Street & \textbf{Average} \\
\midrule
 \multicolumn{1}{l|}{SoM (arXiv'24)~\cite{wang2024shape}} &0.150&0.364&0.290&0.164&0.153&0.119&0.207\\  
 \multicolumn{1}{l|}{SplineGS (CVPR'25)~\cite{park2024splinegs}}& 0.111&0.239&0.220&0.062&0.070&0.055&0.126\\
\midrule
 \multicolumn{1}{l|}{Restormer (CVPR'22)~\cite{zamir2022restormer} + SoM} & 0.099&0.235&0.201&0.130&0.135&0.104&0.151\\  
 \multicolumn{1}{l|}{GShiftNet (CVPR'23)~\cite{li2023simple} + SoM} & 0.089&0.214&0.188&0.107&0.128&0.110&0.139\\  
 \multicolumn{1}{l|}{Restormer (CVPR'22)~\cite{zamir2022restormer} + SplineGS} & 0.063&0.108&0.098&0.036&0.037&0.042&0.064\\  
 \multicolumn{1}{l|}{GShiftNet (CVPR'23)~\cite{li2023simple} + SplineGS} & 0.055&0.108&\second{0.094}&\second{0.029}&\second{0.030}&\best{0.030}&\second{0.058}\\  
\midrule
\multicolumn{1}{l|}{Deblurring 3DGS (ECCV'24)~\cite{lee2023exblurf}} & 0.179&0.205&0.296&0.190&0.210&0.241&0.220\\  
 \multicolumn{1}{l|}{BAD-GS (ECCV'24)~\cite{zhao2024bad}} & 0.060&0.083&0.183&0.030&0.068&0.286&0.119\\  
 \multicolumn{1}{l|}{DyBluRF (CVPR'24)~\cite{sun2024dyblurf}} & \second{0.045}&0.096&0.107&0.037&0.052&0.046&0.064\\  
 \multicolumn{1}{l|}{Deblur4DGS (arXiv'24)~\cite{wu2024deblur4dgs}} & 0.121&0.306&0.384&0.140&0.086&0.064&0.183 \\
 \multicolumn{1}{l|}{MoBluRF (IEEE TPAMI)~\cite{bui2025moblurf}} & 0.059&\second{0.080}&0.102&0.087&0.066&0.046&0.073\\
 \multicolumn{1}{l|}{\textbf{MoBGS (Ours)}} & \best{0.034}&\best{0.052}&\best{0.081}&\best{0.018}&\best{0.025}&\second{0.033}&\best{0.040} \\
\bottomrule
\end{tabular}
}
\label{table:perscene_lpips_deblurring}
\end{center}
\end{table*}

%% file: table/perscene_musiq_deblurring.tex
\begin{table*}
\begin{center}
\caption{\textbf{Per-scene MUSIQ of \textit{full region} for deblurring effect evaluation on the Stereo Blur dataset.} }
\setlength\tabcolsep{6pt} 
\renewcommand{\arraystretch}{1.1}
\scalebox{0.85}{
\begin{tabular}{ c | c | c | c | c |c | c | c }
\toprule
  Methods & Basketball & Children & Sailor & Seesaw & Skating & Street & \textbf{Average} \\
\midrule
 \multicolumn{1}{l|}{SoM (arXiv'24)~\cite{wang2024shape}} & 31.61&49.69&44.62&46.28&54.98&52.11&46.55\\  
 \multicolumn{1}{l|}{SplineGS (CVPR'25)~\cite{park2024splinegs}}& 41.23&39.22&33.09&49.13&46.42&46.52&42.60\\  
\midrule
 \multicolumn{1}{l|}{Restormer (CVPR'22)~\cite{zamir2022restormer} + SoM} & 46.85&\second{60.11}&44.31&51.54&\second{56.55}&\second{56.25}&52.60\\  
 \multicolumn{1}{l|}{GShiftNet (CVPR'23)~\cite{li2023simple} + SoM} & 48.68&59.94&47.38&53.18&\best{57.15}&\best{58.88}&54.20\\  
 \multicolumn{1}{l|}{Restormer (CVPR'22)~\cite{zamir2022restormer} + SplineGS} & 53.46&54.37&\second{52.89}&57.59&50.37&53.28&53.66\\  
 \multicolumn{1}{l|}{GShiftNet (CVPR'23)~\cite{li2023simple} + SplineGS} & \second{54.65}&54.31&51.27&\second{61.39}&54.14&55.67&\second{55.24}\\  
\midrule
\multicolumn{1}{l|}{Deblurring 3DGS (ECCV'24)~\cite{lee2023exblurf}} & 48.39&55.17&41.99&61.20&53.93&44.63&50.88\\  
 \multicolumn{1}{l|}{BAD-GS (ECCV'24)~\cite{zhao2024bad}} & 53.10&58.57&55.82&57.18&46.28&49.39&53.39\\  
 \multicolumn{1}{l|}{DyBluRF (CVPR'24)~\cite{sun2024dyblurf}} & 55.06&53.74&48.95&56.87&45.92&52.44&52.16\\  
 \multicolumn{1}{l|}{Deblur4DGS (arXiv'24)~\cite{wu2024deblur4dgs}} & 48.52&41.13&49.80&49.47&50.01&54.71&48.94 \\
 \multicolumn{1}{l|}{MoBluRF (IEEE TPAMI)~\cite{bui2025moblurf}} & 52.91&55.31&50.09&46.80&46.66&48.25&49.99 \\
 \multicolumn{1}{l|}{\textbf{MoBGS (Ours)}} & \best{57.78}&\best{60.96}&\best{56.47}&\best{61.73}&51.67&55.45&\best{57.34} \\
\bottomrule
\end{tabular}
}
\label{table:perscene_musiq_deblurring}
\end{center}
\end{table*}

%% file: table/perscene_tof_deblurring.tex
\begin{table*}
\begin{center}
\caption{\textbf{Per-scene tOF of \textit{full region} for deblurring effect evaluation on the Stereo Blur dataset.} }
\setlength\tabcolsep{6pt} 
\renewcommand{\arraystretch}{1.1}
\scalebox{0.85}{
\begin{tabular}{ c | c | c | c | c |c | c | c }
\toprule
  Methods & Basketball & Children & Sailor & Seesaw & Skating & Street & \textbf{Average} \\
\midrule
 \multicolumn{1}{l|}{SoM (arXiv'24)~\cite{wang2024shape}} & 0.778&6.201&3.019&1.408&1.257&0.948&2.268\\  
 \multicolumn{1}{l|}{SplineGS (CVPR'25)~\cite{park2024splinegs}}& 0.592&4.152&1.595&0.432&0.418&0.424&1.269\\  
\midrule
 \multicolumn{1}{l|}{Restormer (CVPR'22)~\cite{zamir2022restormer} + SoM} & 0.713&4.810&2.690&1.208&1.145&0.864&1.905\\  
 \multicolumn{1}{l|}{GShiftNet (CVPR'23)~\cite{li2023simple} + SoM} & 0.692&4.952&2.457&1.002&1.149&0.955&1.868\\  
 \multicolumn{1}{l|}{Restormer (CVPR'22)~\cite{zamir2022restormer} + SplineGS} & 0.521&2.084&1.085&0.334&0.299&0.401&0.787\\  
 \multicolumn{1}{l|}{GShiftNet (CVPR'23)~\cite{li2023simple} + SplineGS} & \second{0.478}&1.501&\second{0.948}&\second{0.291}&\best{0.266}&\second{0.359}&\second{0.641}\\  
\midrule
\multicolumn{1}{l|}{Deblurring 3DGS (ECCV'24)~\cite{lee2023exblurf}} & 0.969&2.199&4.893&0.727&1.624&3.158&2.262\\  
 \multicolumn{1}{l|}{BAD-GS (ECCV'24)~\cite{zhao2024bad}} & 0.499&\second{1.144}&3.283&0.329&0.780&3.128&1.527\\  
 \multicolumn{1}{l|}{DyBluRF (CVPR'24)~\cite{sun2024dyblurf}} &0.481&1.354&1.190&0.465&0.518&0.433&0.740\\  
 \multicolumn{1}{l|}{Deblur4DGS (arXiv'24)~\cite{wu2024deblur4dgs}} & 0.706&4.678&4.428&0.990&0.597&0.480&1.980 \\
 \multicolumn{1}{l|}{MoBluRF (IEEE TPAMI)~\cite{bui2025moblurf}} & 0.487&1.226&1.172&0.600&0.561&0.406&0.742 \\
 \multicolumn{1}{l|}{\textbf{MoBGS (Ours)}} & \best{0.289}&\best{0.761}&\best{0.848}&\best{0.197}&\second{0.272}&\best{0.357}&\best{0.454} \\
\bottomrule
\end{tabular}
}
\label{table:perscene_tof_deblurring}
\end{center}
\end{table*}

%% file: table/perscene_psnr_deblurring.tex
\begin{table*}
\begin{center}
\caption{\textbf{Per-scene PSNR of \textit{full region} for deblurring effect evaluation on the Stereo Blur dataset.} }
\setlength\tabcolsep{6pt} 
\renewcommand{\arraystretch}{1.1}
\scalebox{0.85}{
\begin{tabular}{ c | c | c | c | c |c | c | c }
\toprule
  Methods & Basketball & Children & Sailor & Seesaw & Skating & Street & \textbf{Average} \\
\midrule
 \multicolumn{1}{l|}{SoM (arXiv'24)~\cite{wang2024shape}} & 24.58&22.88&22.82&23.78&26.04&27.05&24.53\\  
 \multicolumn{1}{l|}{SplineGS (CVPR'25)~\cite{park2024splinegs}}&27.07&25.44&26.23&31.00&33.91&31.67&29.22\\
\midrule
 \multicolumn{1}{l|}{Restormer (CVPR'22)~\cite{zamir2022restormer} + SoM} & 24.99&22.83&23.13&24.23&25.79&27.24&24.70\\  
 \multicolumn{1}{l|}{GShiftNet (CVPR'23)~\cite{li2023simple} + SoM} & 25.14&22.74&22.99&24.56&25.15&26.09&24.45\\  
 \multicolumn{1}{l|}{Restormer (CVPR'22)~\cite{zamir2022restormer} + SplineGS} & 28.27&27.13&\second{27.90}&32.03&\second{34.91}&31.61&\second{30.31}\\  
 \multicolumn{1}{l|}{GShiftNet (CVPR'23)~\cite{li2023simple} + SplineGS} & 28.38&26.49&27.79&\second{32.17}&34.56&\best{32.45}&\second{30.31}\\ 
\midrule
\multicolumn{1}{l|}{Deblurring 3DGS (ECCV'24)~\cite{lee2023exblurf}} & 18.89&20.21&18.36&23.99&22.76&18.61&20.47\\  
 \multicolumn{1}{l|}{BAD-GS (ECCV'24)~\cite{zhao2024bad}} & \second{29.09}&24.56&19.18&30.92&29.63&18.22&25.27\\  
 \multicolumn{1}{l|}{DyBluRF (CVPR'24)~\cite{sun2024dyblurf}} & 27.62&\second{28.00}&26.16&31.41&31.91&29.81&29.15\\  
 \multicolumn{1}{l|}{Deblur4DGS (arXiv'24)~\cite{wu2024deblur4dgs}} & 22.91&21.58&19.07&24.53&27.64&26.48&23.70 \\
 \multicolumn{1}{l|}{MoBluRF (IEEE TPAMI)~\cite{bui2025moblurf}} & 26.56&27.75&26.91&29.66&32.79&\second{31.97}&29.27 \\
 \multicolumn{1}{l|}{\textbf{MoBGS (Ours)}} & \best{29.83}&\best{28.18}&\best{28.51}&\best{32.57}&\best{35.32}&31.34&\best{30.96} \\
\bottomrule
\end{tabular}
}
\label{table:perscene_psnr_deblurring}
\end{center}
\end{table*}